\documentclass[twocolumn,journal]{IEEEtran}
\usepackage[T1]{fontenc}
\usepackage[latin9]{inputenc}
\usepackage{array}
\usepackage{verbatim}
\usepackage{float}
\usepackage{calc}
\usepackage{multirow}
\usepackage{amsmath}
\usepackage{amssymb}
\usepackage{graphicx}

\makeatletter

\providecommand{\tabularnewline}{\\}
\floatstyle{ruled}
\newfloat{algorithm}{tbp}{loa}
\providecommand{\algorithmname}{Algorithm}
\floatname{algorithm}{\protect\algorithmname}

\usepackage[caption=false,font=footnotesize]{subfig}

\@ifundefined{showcaptionsetup}{}{%
 \PassOptionsToPackage{caption=false}{subfig}}
\usepackage{subfig}
\makeatother

\begin{document}

\title{Center of Gravity-based Approach for Modeling Dynamics of Multisection
Continuum Arms$^*$}

\author{Isuru~S.~Godage$^{*}$,~\IEEEmembership{Member,~IEEE,} Robert
J. Webster III$^{\dagger}$,~\IEEEmembership{Senior~Member,~IEEE,} and
\\Ian D. Walker$^{\ddagger}$,~\IEEEmembership{Fellow,~IEEE,} }
\maketitle
\begin{abstract}
Multisection continuum arms offer complementary characteristics to
those of traditional rigid-bodied robots. Inspired by biological appendages,
such as elephant trunks and octopus arms, these robots trade rigidity
for compliance, accuracy for safety, and therefore exhibit strong
potential for applications in human-occupied spaces. Prior work has
demonstrated their superiority in operation in congested spaces and
manipulation of irregularly-shaped objects. However, they are yet
to be widely applied outside laboratory spaces. One key reason is
that, due to compliance, they are difficult to control. Sophisticated
and numerically efficient dynamic models are a necessity to implement
dynamic control. In this paper, we propose a novel, numerically stable,
center of gravity-based dynamic model for variable-length multisection
continuum arms. The model can accommodate continuum robots having
any number of sections with varying physical dimensions. The dynamic
algorithm is of $\mathcal{O}\left(n^{2}\right)$ complexity, runs
at 9.5~kHz, simulates 6-8 times faster than real-time for a three-section
continuum robot, and therefore is ideally suited for real-time control
implementations. The model accuracy is validated numerically against
an integral-dynamic model proposed by the authors and experimentally
for a three-section, pneumatically actuated variable-length multisection
continuum arm. This is the first sub real-time dynamic model based
on a smooth continuous deformation model for variable-length multisection
continuum arms.
\end{abstract}

\begin{IEEEkeywords}
continuum arms, dynamics, center of gravity, real-time

\end{IEEEkeywords}

\section{Introduction\label{sec:Introduction}}

\begin{table}[b]
$*$ School of Computing, DePaul University, Chicago, IL 60604. email: {igodage@depaul.edu}. $\dagger$
Dept. of Mechanical Engineering, Vanderbilt University, Nashville,
TN 37212. $\ddagger$ Dept. of Electrical and Computer Engineering,
Clemson University, SC 29634. \vspace{1mm} \\ This work is supported
in part by the National Science Foundation grant IIS-1718755. $*$ Submitted to IEEE Transactions on Robotics.
\end{table}
\IEEEPARstart{R}{igid}-bodied robots have been the backbone of the
robotic industrial revolution which has not only significantly improved
throughput but also relieved humans of most of the mundane, repetitive,
dangerous, and dirty tasks of assembly lines. Rigid-linked industrial
robots have high payload capacity and precision superior to human
capabilities. However, the lack of compliance of rigid robots renders
them dangerous and therefore industrial robot task-spaces are often
restricted of human presence. In addition, due to the structural rigidity,
they are poorly adaptable to environmental interaction and yield poor
performance in unstructured environments \cite{camarillo2008mechanics}.
There is currently great interest in robots that work cooperatively
with humans \cite{Wilson1993}, which implies a need for inherently
human-safe robotic manipulators. Continuum robots have been proposed
as a potential solution to serve niche applications where adaptability,
compliance, and human safety are critical \cite{qi2017design}. In
this paper, we refer to continuum robots as those robotic structures
that lack rigid frames and generate motion through smooth, continuous
structural deformation, such as the robots reported in \cite{ivanescu2005variable,Mahl2014variable,Renda2014dynamic,Cheng2012design,cianchetti2013stiff,kim2012design}.

\begin{figure}[t]
\begin{centering}
\subfloat[]{\begin{raggedright}
\includegraphics[height=1.4in]{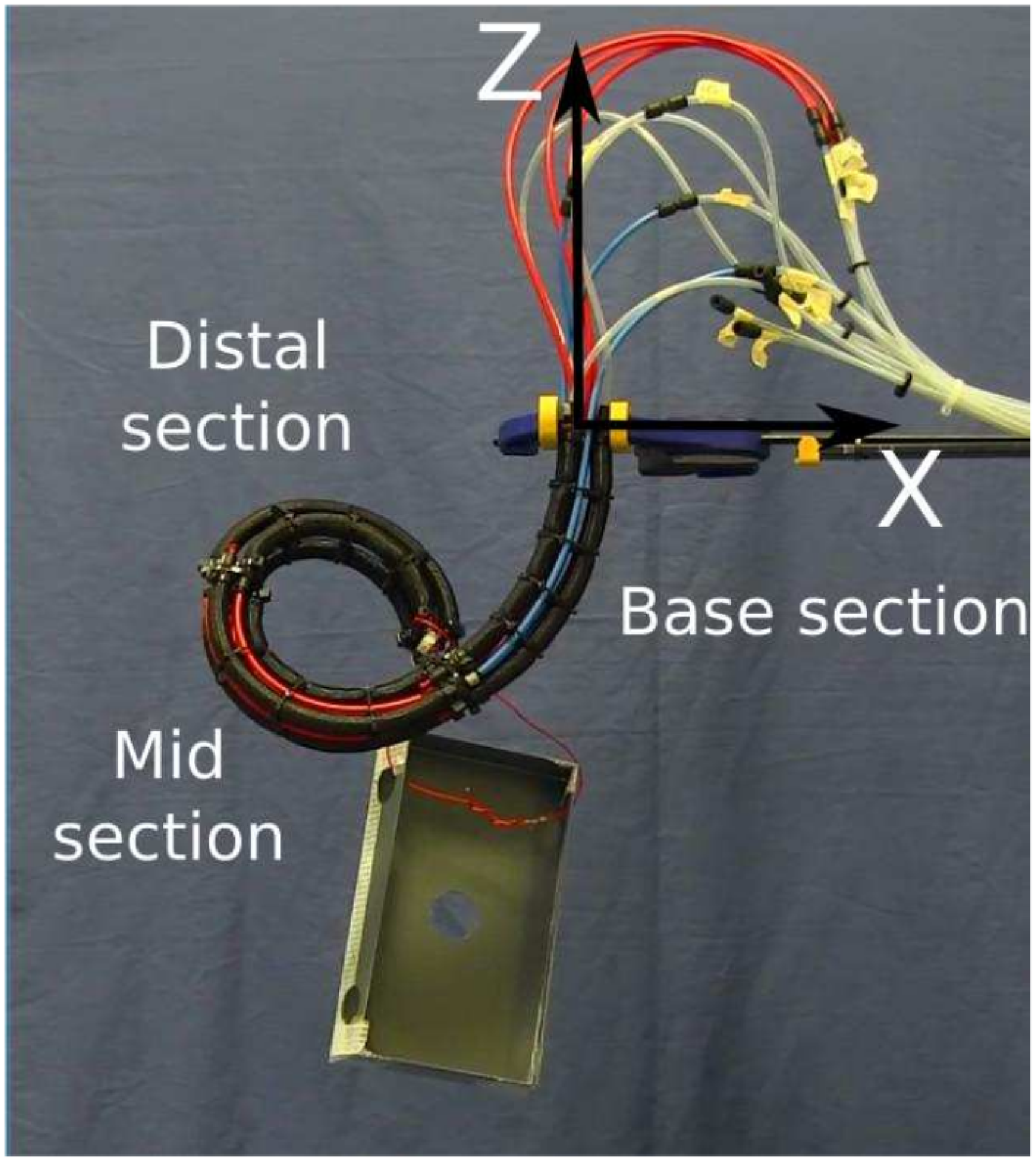}
\par\end{raggedright}
\label{fig:iitArm}}\subfloat[]{\begin{raggedright}
$\!\!\!\!$\includegraphics[height=1.4in]{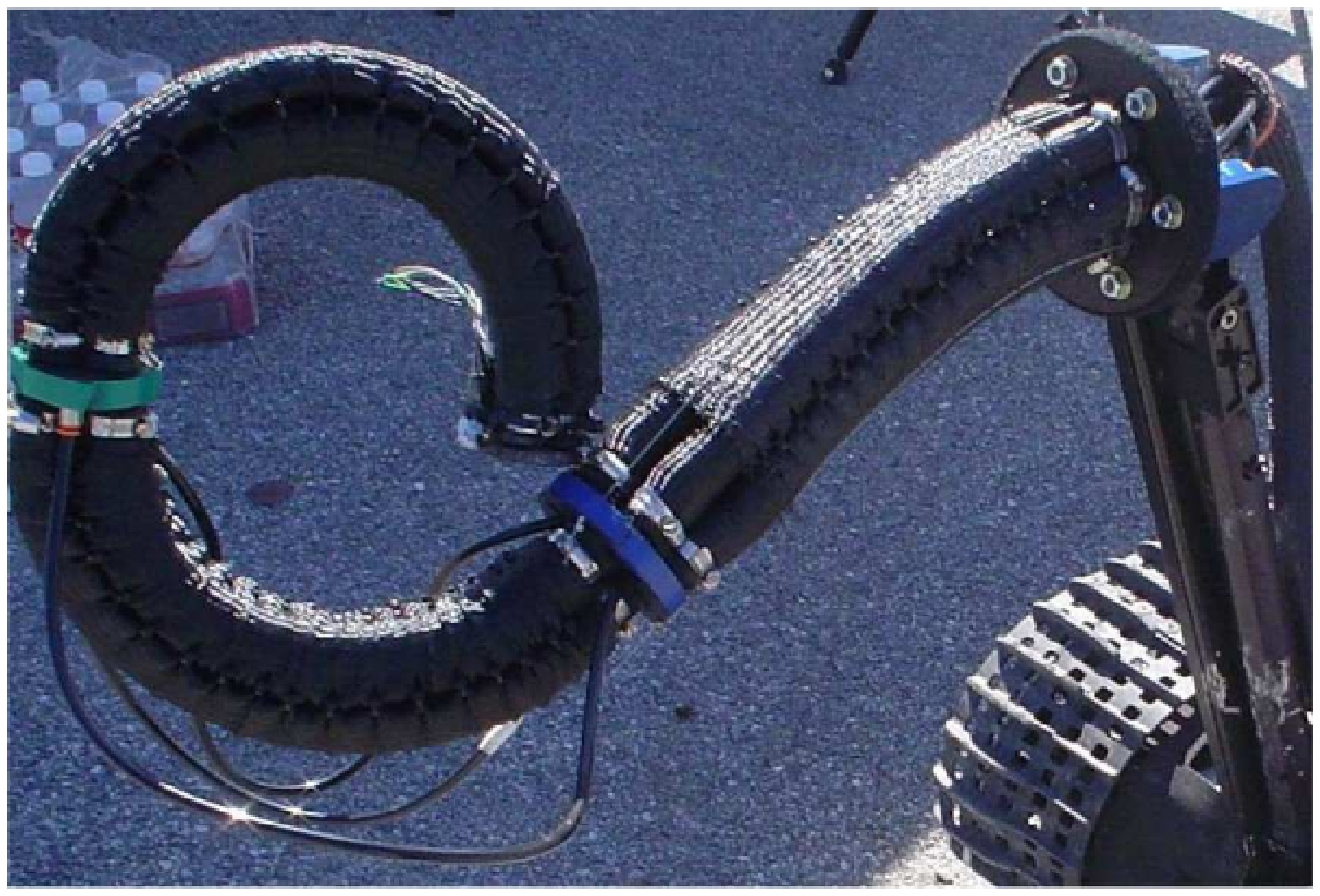}
\par\end{raggedright}
\label{fig:octarm}}
\par\end{centering}
\caption{Examples of pneumatic muscle actuator (PMA) powered variable-length
multisection continuum arms. (a) the continuum arm developed at the
Italian Institute of Technology is used to validate the dynamic model
proposed in this paper \cite{godage2015modal}, (b) OctArm-IV \cite{grissom2006design}
continuum manipulator developed at Clemson University.}
\end{figure}

Continuum arms are inspired by biological appendages such as elephant
trunks and octopus arms. Made entirely out of muscle fibers, continuum
arms structures are highly deformable to achieve complex geometrical
shapes. Despite being made entirely out of muscles, they demonstrate
compelling benchmarks in terms of forces and precision of operation
\cite{cieslak1999elephant,Neppalli2007OctArm,zheng2013model,laschi2012soft,cianchetti2012design}.
Often constructed from elastic material, continuum arms aim to imitate
such behavior by generating complex smooth geometric shapes through
structural deformation. The smaller continuum robots target operation
in smaller spaces such as inside human bodies during minimally invasive
surgeries \cite{burgner2015continuum}, and are actuated by elastic
tubes or tendons. The larger variants, constructed to handle macro
or human body scale objects are often powered by pneumatic muscle
actuators (PMA). PMA's, also known as McKibben actuators have number
of desirable features, such as ease of design, fabrication, and high
power-to-weight ratio, and therefore are sought after in continuum
arm designs. In this paper we focus on PMA powered variable-length
multisection continuum arms. There are several key features common
to this type of manipulator. Unlike tendon-actuated continuum arms,
they are fabricated by serially stacking continuum sections where
each continuum section consists of multiple PMAs (typically three,
though four actuators are also possible \cite{Renda2014dynamic})
and are capable of generating omindirectional bending deformation
independent of other sections. Since there are no backbones, continuum
sections undergo axial length changes, extend or contract, depending
on the PMA operation mode. %
Figures \ref{fig:iitArm} and \ref{fig:octarm} show a couple of variable-length
multisection continuum arm prototypes. Due to their unique mechanical
characteristics, deriving mathematical models for these robots has
been a challenge.%

\subsection{Prior Work on Dynamic Modeling of Continuum Arms\label{subsec:Prior-Work-on}}

Early continuum-style (which are not truly continuum without continuously
bending deformation) robots have been discretized rigid structures
\cite{chirikjian1995kinematically,Hannan2001The} that mimicked smooth
bending. %
The computational constraints that prevailed at the time motivated
numerically efficient parametric or modal approaches \cite{chirikjian1995kinematics}.%
{} However, such low dimensional methods did not fully capture the complete
task-space and suffered from numerical instabilities \cite{chirikjian1994modal}.
Other early continuum-style robots and discrete-link dynamic models
include \cite{zheng2013model,kang2013design,Mahl2014variable}.%
Cosserat rod theory has been proposed to model quasi-statics of tendon
actuated inextensible flexible backbone and concentric continuum robots
\cite{jones2009three,rucker2011statics,Renda2014dynamic}. The work
in \cite{Renda2014dynamic} employed a similar approach to model the
dynamics of a multibending soft manipulator but, owing to the complexity
associated with highly deformable bodies, reported inefficient simulation
times. The work reported in \cite{xu2010analytic} utilized elliptic
integrals to develop kinematics and statics of miniature single continuum
section. Dynamics based on Kane's method was reported in \cite{Rone2014continuum}
to model the dynamics of a tendon-actuated continuum manipulator.
The work in \cite{trivedi2008geometrically} proposed and validated
a planar and static Cosserat rod model for PMA actuated variable-length
multisection continuum sections, but the approach has not been extended
for modeling spatial dynamics.

Another avenue to derive equations of motion (EoM) is to utilize energy-based
methods such as the Lagrangian formulation. During operation, the
relative displacement between points of a continuum body varies and
thus limits the use of numerically efficient algorithms \cite{featherstone2008rigid}.
Theoretical models for inextensible, unidimensional, rope-like mechanisms
were proposed in \cite{mochiyama2006hyper}, but continuum arms have
multiple degrees of freedom (DoF). %

The kinematic model reported in \cite{jones2006kinematics} %
laid the foundation for curve parametric models for variable-length
continuum arms%
. Nonetheless, the use of circular arc parameters resulted in complex
nonlinear terms and numerical instabilities for straight-arm poses
to limit the model's extensibility for modeling dynamics. For an in-depth
treatment of the limitations of curve parametric models%
, see \cite{godage2015modal}. An energy-based derivation of planar
dynamic models for OctArm variable-length continuum manipulator \cite{grissom2006design}
were reported in \cite{tatlicioglu2007dynamic,tatlicioglu2007new}%
. However, continuum arms are capable of spatial operation, and the
models were not experimentally validated. In addition, the resulting
EoM were nonlinear, complex, and of integral nature, and therefore
numerically inefficient and unstable. %

Prior work by the first author proposed a %
modal method to overcome the numerical instabilities and inefficiencies
present in curve parametric models. Therein, the terms of the homogeneous
transformation matrix (HTM) of continuum sections were approximated
by multivariate polynomials \cite{godage2011novel,godage2015modal}
where the degree of polynomials could be chosen to meet desired error
metrics. %
The model laid the foundation for formulating EoM of variable-length
continuum sections \cite{godage2011shape,godage2011dynamics,godage2012pneumatic}.
The extended recursive formulation was later validated for a variable-length
multisection continuum manipulator \cite{godage2016dynamics}\textbf{.}
Therein, the integral terms are presolved %
to improve the numerical performance.%

Numerically efficient (via rigid body dynamic algorithms) lumped models
have also been applied for continuum robots. However, such models
require a large number of discrete joints to %
approximate the deformation%
{} \cite{kang2012dynamic,kang2013design,zheng2013model}. %
Some work has attempted to trade numerical efficiency for modeling
accuracy by using relatively few rigid segments \cite{Giri2011three,Khalil2007dynamic}. 

The key motivation of this paper is to introduce a lumped model without
betraying the continuous nature of the resulting expressions. Our
prior work introduced a center of gravity (CoG) based modeling approach
for a single continuum section \cite{godage2015efficient,godage2015accurate}.
Therein, the EoM were derived for a point mass at the CoG of the continuum
section. %
Thus, instead of an integral formulation, the process resulted in
a compact model and superior numerical efficiency. In the derivation
process, due to the physical dimensions of the robot, we did not consider
the angular kinetic energy as the energy contribution was less than
3\%. But this will not be the case for all continuum arms. In addition,
the model was limited to a single continuum section where continuum
arms with multiple sections are required for performing useful tasks
such as whole arm manipulation \cite{li2011determining}and spatial
trajectory tracking \cite{godage2011novel}.

\subsection{Contribution\label{subsec:Contribution}}

In this work, we extend and generalize our CoG-based spatial dynamic
model derived for a single continuum section \cite{godage2015accurate},
evaluate against the integral dynamics proposed in \cite{godage2016dynamics}
to verify the numerical accuracy and computational efficiency, and
validate the model against spatial dynamic responses of the prototype
arm shown in Fig. \ref{fig:iitArm}. Beyond our prior work reported
in \cite{godage2016dynamics,godage2015accurate,godage2015efficient},
the proposed dynamic model; (1) accommodates variable-length multisection
continuum arms with arbitrary number of sections and a wide range
of length and radii combinations, (2) considers both linear and angular
kinetic energies of the continuum arm at the CoG for better system
energy accuracy, (3) achieves energy matching via a series of energy
shaping coefficients that are constant for any variable-length multisection
continuum arms, (4) employs the results from \cite{godage2016dynamics}
to systematically derive the EoM terms recursively, (5) demonstrates
$\mathcal{O}\left(n^{2}\right)$ complexity for the first time for
a dynamic model based on continuous (non-discretized) deformation
representation, and for a three-section continuum arm, (6) runs at
9.5~kHz (step execution rate), and (7) achieves sub real-time dynamic
simulation in Matlab Simulink environment. Therefore the proposed
model unifies the ideas of lumped parametric approaches of discrete
rigid-bodied robotics and continuous (integral) approaches of continuum
robotics and is expected to lay a strong numerical and algorithmic
foundation for implementing dynamic control schemes. %

\section{Kinematics of Centers of Gravity\label{sec:Kinematics-of-Centers}}

\subsection{System Model and Assumptions\label{subsec:System-Model-and}}

Tables \ref{tab:nom_symb} and \ref{tab:nom_op} list the nomenclatures
of mathematical symbols and operators employed in this paper. Figure
\ref{fig:schematicArm} shows the schematic of a multisection continuum
arm with $n\in\mathbb{Z}^{+}$ sections. %
The sections are numbered starting from the base continuum section
(index 1) attached to the task-space coordinate system, $\left\{ O\right\} $.
Any $i^{th}$ continuum section (Fig. \ref{fig:section_slice_vels_forces})
is assumed to be actuated by three extending PMAs which are mounted
on plates situated at either end at $r_{i}\in\mathbb{R}^{+}$ distance
from the neutral axis and $\frac{2\pi}{3}$ rad apart. Let the unactuated
length of PMAs be $L\in\mathbb{R}^{+}$ , the maximum length change
$l_{max}$, and the joint-space vector of the continuum section, $\boldsymbol{q}_{i}=\left[l_{i1},l_{i2},l_{i3}\right]^{T}$
where $l_{ij}\in\left[0,l_{max}\right]$ $\forall j\in\left\{ 1,2,3\right\} $.
The joint where the $\left(i+1\right)^{th}$ continuum section is
attached introduces $\sigma_{i}\in\mathbb{R}_{0}^{+}$ linear displacement
along and $\gamma_{i}\in\mathbb{R}_{0}$ angular displacement about
the +Z axis of $\left\{ O_{i}\right\} $. As the PMAs are constrained
to maintain $r_{i}$ clearance normal to the neutral axis, differential
length changes cause the section to bend or extend (when length changes
are equal) \cite{grissom2006design}. The subsequent derivations rely
on the assumptions that the continuum sections bend in circular arc
shapes, have constant mass $m_{i}\in\mathbb{R}^{+}$, and uniform
linear density%
\footnote{These are reasonable assumptions under typical operating conditions
without large external forces as shown in \cite{godage2015modal}
and \cite{godage2016dynamics}.}.%
\begin{center}
\begin{table}[tb]
\begin{raggedright}
\caption{Nomenclature of Mathematical Symbols\label{tab:nom_symb}}
\par\end{raggedright}
\footnotesize
\begin{raggedright}
\begin{tabular}{>{\raggedleft}p{0.15\columnwidth}|>{\raggedright}p{0.75\columnwidth}}
\hline 
\textbf{Symbol} & \textbf{Definition}\tabularnewline
\hline 
$i$ & Continuum section index$^{\sharp}$. %
\tabularnewline
$\overline{[\,]}$ & Refers to the center of gravity-related terms\tabularnewline
$r_{i}$, $L_{i}$,$l_{ij}$  & Radius, original length, and $j^{th}$ actuator length change\tabularnewline
$\boldsymbol{q}$,$\boldsymbol{q}_{i}$,$\boldsymbol{q}^{i}$ & Complete, $i^{th}$, and up to $i^{th}$ section joint space vector$^{*}$\tabularnewline
$\!\!\!\!\!\!$$\left\{ O\right\} $,$\left\{ \!O_{i}\!\right\} $,$\left\{ \!O_{i}'\!\right\} $ & Task, base, and moving coordinate frames\tabularnewline
$\mathbf{T}_{i}$,$\boldsymbol{p}_{i}$,$\mathbf{R}_{i}$ & HTM$^{\ddagger}$, position, and rotation matrices relative to $\left\{ O_{i}\right\} $\tabularnewline
$\mathbf{T}^{i}$,$\boldsymbol{p}^{i}$,$\mathbf{R}^{i}$ & HTM, position, and rotation matrices relative to $\left\{ O\right\} $\tabularnewline
$\xi_{i}$ & Scalar to define $\left\{ \!O_{i}'\!\right\} $ along the continuum
section\tabularnewline
$m_{i}$ & Mass of continuum section\tabularnewline
$\mathcal{K}$, $\mathcal{K}_{i}$ & Total and $i^{th}$ section kinetic energy\tabularnewline
$\mathcal{K}_{i}^{\omega}$, $\mathcal{K}_{i}^{\upsilon}$ & Angular and linear kinetic energies of continuum section\tabularnewline
$\mathcal{P}$, $\mathcal{P}_{i}$ & Total and $i^{th}$ section potential energy\tabularnewline
$\mathcal{M}_{i}^{\upsilon}$, $\mathcal{M}_{i}^{\omega}\!\!$ & Disc linear and angular inertia matrices\tabularnewline
$\mathbf{M}$, $\mathbf{C}$ & Complete inertia and Coriolis/Centrifugal matrices\tabularnewline
$\mathbf{M}_{i}^{\upsilon}$, $\mathbf{M}_{i}^{\omega}\!\!$ & Generalized linear and angular inertia matrices\tabularnewline
$\mathbf{C}_{i}^{\upsilon}$, $\mathbf{C}_{i}^{\omega}$ & Linear, angular Coriolis/Centrifugal force matrices\tabularnewline
$\boldsymbol{G}$,$\boldsymbol{G}_{i}$ & Complete, $i^{th}$ section conservative force vectors\tabularnewline
$\mathbf{J}_{i}^{\upsilon}$,$\mathbf{H}_{i}^{\upsilon}$ & Linear velocity Jacobian and Hessian w.r.t to $\left\{ \!O_{i}'\!\right\} $\tabularnewline
$\mathbf{J}_{i}^{\Omega}$,$\mathbf{H}_{i}^{\Omega}$ & Angular velocity Jacobian and Hessian w.r.t to $\left\{ \!O_{i}'\!\right\} $\tabularnewline
$\mathbf{K}_{i}^{e}$ & Elastic stiffness coefficient matrix \tabularnewline
$\boldsymbol{\tau}_{e}$ & Complete input force vector in the joint space\tabularnewline
$\mathbf{I}_{3}$ & Rank 3 identity matrix\tabularnewline
\hline 
\end{tabular}
\par\end{raggedright}
\vspace{1mm}

$\sharp$ Subscript $i$ represents the $i^{th}$ continuum section
parameters whereas superscript stands for terms associated with up
to the $i^{th}$ continuum section.

$*$Lowercase, boldface italics (i.e., $\boldsymbol{q}_{j}$) denote
vectors and regular lowercase italics (i.e., $l_{ij}$or $h$) denote
vector/matrix elements or constants. Matrices are denoted by boldface
uppercase letters (i.e., $\mathbf{T}$, $\mathbf{M}_{i}^{\omega}$)

$\ddagger$ Homogeneous transformation matrix (HTM).

All quantities are represented in metric units.
\end{table}
\par\end{center}

\begin{center}
\begin{table}[tb]
\begin{raggedright}
\caption{Nomenclature of Mathematical Operators\label{tab:nom_op}}
\par\end{raggedright}
\begin{raggedright}
\footnotesize%
\begin{tabular}{>{\raggedleft}p{0.12\columnwidth}|>{\raggedright}p{0.76\columnwidth}}
\hline 
\textbf{Operator} & \textbf{Definition}\tabularnewline
\hline 
$\left(\quad\right)_{,\boldsymbol{q}}$ & Partial derivative with respect to elements of $\boldsymbol{q}$ along
the dimension of $\boldsymbol{q}$. Eg. if $\boldsymbol{q}\in\mathbb{R}^{n\times1}$
and $\mathbf{A}\in\mathbb{R}^{u\times v}$, then $\mathbf{A}_{,\boldsymbol{q}}\in\mathbb{R}^{nu\times v}$
and $\mathbf{A}_{,\boldsymbol{q}^{T}}\in\mathbb{R}^{u\times nv}$
respectively.\tabularnewline
$\left(\quad\right)^{\vee}$ & Forms the velocity vector from skew-symmetric angular velocity matrix\tabularnewline
${\textstyle \int}$ & Integration from 0 to 1 with respect to $\xi_{i}$\tabularnewline
 $\mathbb{T}_{2}$ & Trace operator (involving only the first two diagonal elements) on
a $3\times3$ matrix or sub-matrix\tabularnewline
\hline 
\end{tabular}
\par\end{raggedright}
\vspace{2mm}

\footnotesize%
\end{table}
\par\end{center}

\begin{center}
\begin{figure}[tb]
\begin{centering}
\subfloat[]{\begin{centering}
\includegraphics[height=0.37\columnwidth]{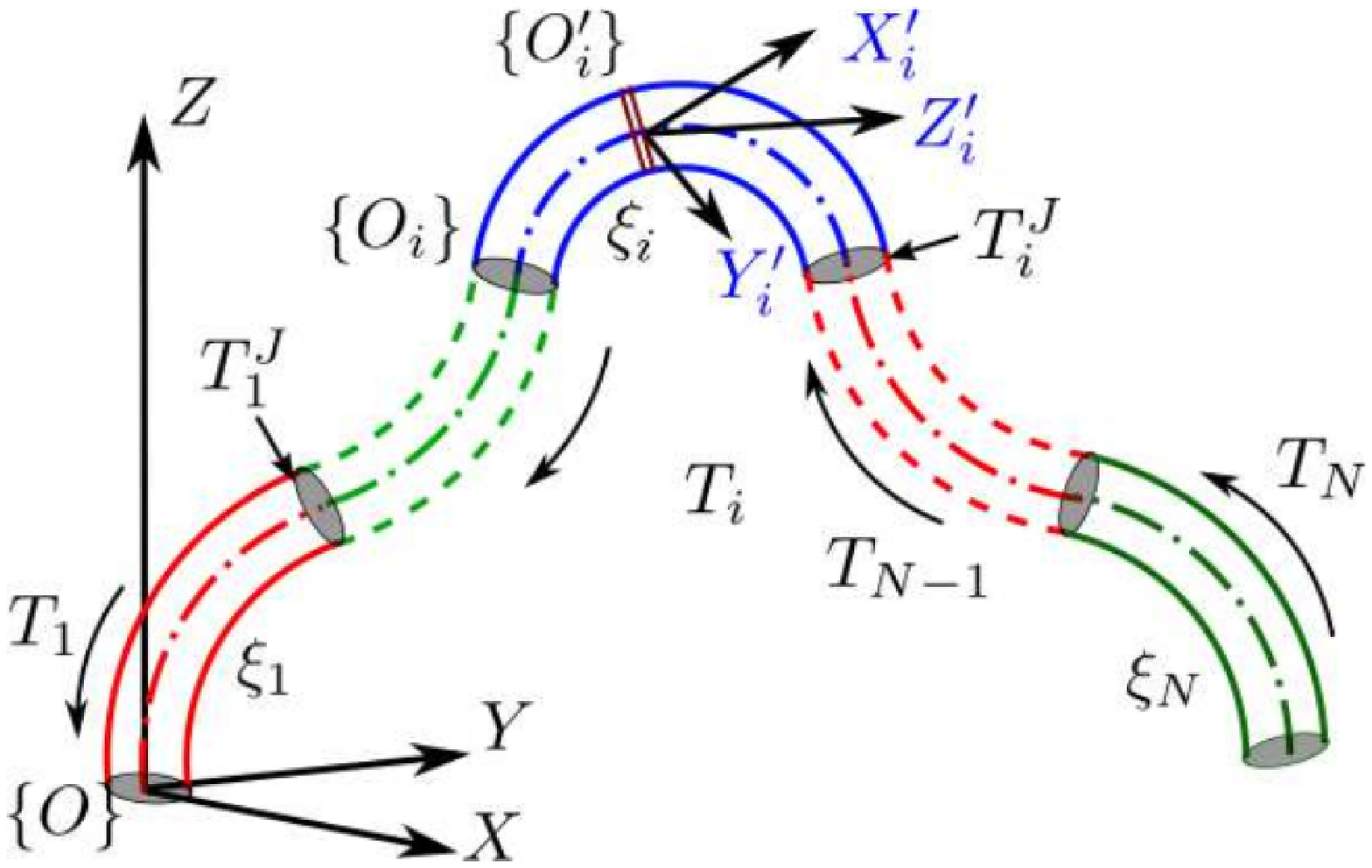}
\par\end{centering}
\label{fig:schematicArm}}\subfloat[]{\begin{centering}
\includegraphics[height=0.37\columnwidth]{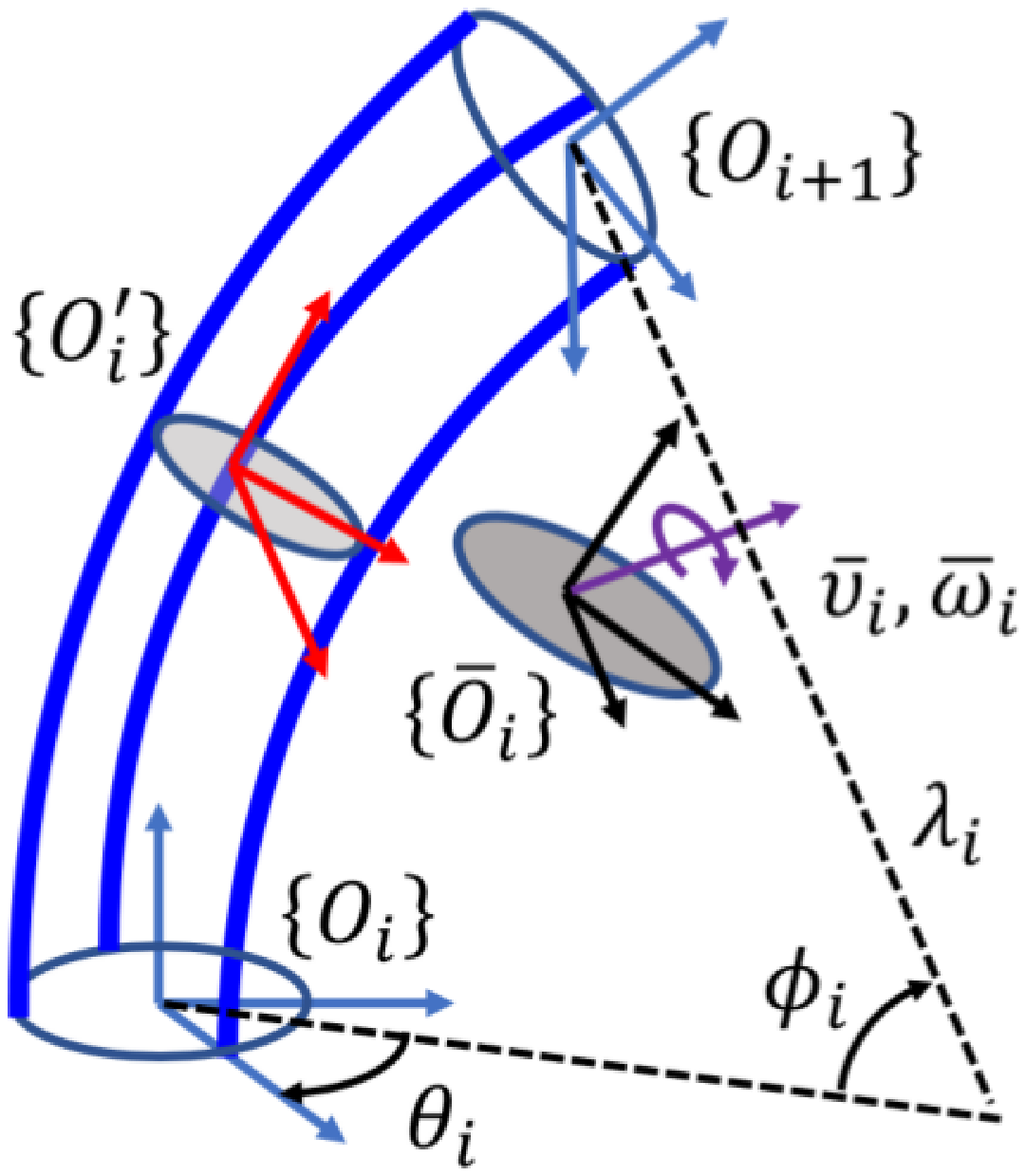}
\par\end{centering}
\label{fig:section_slice_vels_forces}}
\par\end{centering}
\caption{(a) Schematic of a multisection continuum arm. (b) Schematic of an
infinitesimally thin slice the CoG of any $i^{th}$ continuum section.}
\end{figure}
\par\end{center}

\vspace{-25mm}

\subsection{Recursive Velocities, Jacobians, and Hessians\label{subsec:Recursive-Velocities,-Jacobians,}}

The kinematics of continuum arms has been well studied over the years
\cite{burgner2015continuum,jones2006kinematics,webster2010design,godage2015dual}.
The proposed work uses the modal kinematics \cite{godage2015modal}
for subsequent derivation of the EoM. This section provides a review
of the modal kinematics for multisection continuum arms. A detailed
exposition of kinematics is found in \cite{godage2015modal}.

The deformation of a continuum section can be defined by the curve
parameters $\lambda\left(\boldsymbol{q}_{i}\right)\in\mathbb{R}^{+}$
radius of the circular arc, $\phi\left(\boldsymbol{q}_{i}\right)\in\mathbb{R}_{0}^{+}$
angle subtended by the circular arc, and $\theta\left(\boldsymbol{q}_{i}\right)\in\left(-\pi,\pi\right]$\footnote{As shown in \cite{godage2015modal}, the curve parameters are also
functions of unactuated length of PMAs, $L_{i}$, and radius of continuum
section, $r_{i}$, but are not included in the notation (constants
for a given continuum arm) for brevity.} (see Fig. \ref{fig:section_slice_vels_forces}). Employing the curve
parameters, the HTM of $\left\{ O_{i}'\right\} $ along the neutral
axis of the $i^{th}$ continuum section at $\xi_{i}\in\left[0,1\right]$
with respect to $\left\{ O_{i}\right\} $, $\mathbf{T}_{i}:\left(\boldsymbol{q}_{i},\xi_{i}\right)\mapsto\mathbb{SE}^{3}$,
is computed as

\begin{align}
\mathbf{T}_{i} & =\mathbf{R}_{Z}\left(\theta_{i}\right)\mathbf{P}_{X}\left(\lambda_{i}\right)\mathbf{R}_{Y}\left(\xi_{i}\phi_{i}\right)\mathbf{P}_{X}\left(-\lambda_{i}\right)\mathbf{R}_{Z}\left(\theta_{i}\right)\cdots\nonumber \\
 & \qquad\mathbf{P}_{Z}\left(\sigma_{i}\right)\mathbf{R}_{Z}\left(\gamma_{i}\right)=\left[\begin{array}{cc}
\mathbf{R}_{i} & \boldsymbol{p}_{i}\\
\boldsymbol{0} & 1
\end{array}\right]\label{eq:HTM_section}
\end{align}
where $\mathbf{P}_{X}\in\mathbb{SE}^{3},$ $\mathbf{R}_{Z}\in\mathbb{SO}^{3}$,
and $\mathbf{R}_{Y}\in\mathbb{SO}^{3}$ are HTM that denotes translation
along the +X axis, rotation about the +Z and +Y axes respectively.
$\mathbf{R}_{i}\left(\boldsymbol{q}_{i},\xi_{i}\right)\in\mathbb{R}^{3\times3}$
is the resultant rotation matrix and $\boldsymbol{p}_{i}\left(\boldsymbol{q}_{i},\xi_{i}\right)\in\mathbb{R}^{3}$
is the position vector. The scalar $\xi_{i}$ denotes any point along
the neutral axis where $\xi_{i}=0$ is the base where $\left\{ O_{i}'\right\} \equiv\left\{ O_{i}\right\} $
and $\xi_{i}=1$ is the tip of the continuum section. We then apply
$15^{th}$ order multivariate Taylor series expansion on the terms
of \eqref{eq:HTM_section} to obtain numerically efficient and stable
modal form of the HTM (see \cite{godage2015modal}).

Employing the continuum section HTM given in \eqref{eq:HTM_section}
and principles of kinematics of serial robot chains, the HTM of any
$i^{th}$ section with respect to the task-space coordinate system
$\left\{ O\right\} $, $\mathbf{T}^{i}:\left(\boldsymbol{q}^{i},\xi_{i}\right)\mapsto\mathbb{SE}^{3}$,
is given by

\begin{align}
\mathbf{T}^{i} & =\prod_{k=1}^{i}\mathbf{T}_{i}=\left[\begin{array}{cc}
\mathbf{R}^{i} & \boldsymbol{p}^{i}\\
\boldsymbol{0} & 1
\end{array}\right]\label{eq:HTM_ith}
\end{align}
where $\mathbf{R}^{i}\left(\boldsymbol{q}^{i},\xi_{i}\right)\in\mathbb{R}^{3\times3}$
and $\boldsymbol{p}_{i}\left(\boldsymbol{q}^{i},\xi_{i}\right)\in\mathbb{R}^{3}$
define the position and orientation of $\left\{ O_{i}'\right\} $
along the neutral axis at $\xi_{i}$ of the $i^{th}$ continuum section%
. 

The HTM in \eqref{eq:HTM_ith} can be expanded to obtain the recursive
form of the kinematics as

\begin{align}
\begin{split}\mathbf{R}^{i} & =\mathbf{R}^{i-1}\mathbf{R}_{i}\\
\boldsymbol{p}^{i} & =\boldsymbol{p}^{i-1}+\mathbf{R}^{i-1}\boldsymbol{p}_{i}
\end{split}
\label{eq:pR_recursive}
\end{align}
where $\mathbf{R}^{i-1}\left(\boldsymbol{q}^{i-1}\right)\in\mathbb{R}^{3\times3}$
and $\boldsymbol{p}_{i}\left(\boldsymbol{q}^{i-1}\right)\in\mathbb{R}^{3}$
is the section tip rotation matrix and position vector of the preceding
continuum section. Notice the absence of $\xi_{i}$ as $\xi_{k}=1\forall k<i$
as per the definition of $\xi_{i}$ (see \cite{godage2015modal}).
Also, from now on, the dependency variables are not included in the
equations for reasons of brevity. 

Exploiting the integral Lagrangian formulation \cite{godage2016dynamics},
we consider a thin disc at $\xi_{i}$ (which lies on the XY plane
of $\left\{ O_{i}'\right\} $. %
Utilizing \eqref{eq:pR_recursive}, the angular and linear body velocities
with respect to $\left\{ O_{i}'\right\} $, $\boldsymbol{\omega}_{i}\left(\boldsymbol{q}^{i},\dot{\boldsymbol{q}}^{i}\right)\in\mathbb{R}^{3}$
and $\boldsymbol{\upsilon}_{i}\left(\boldsymbol{q}^{i},\dot{\boldsymbol{q}}^{i}\right)\in\mathbb{R}^{3}$
respectively, can be defined as

\begin{align}
\begin{split}\boldsymbol{\Omega}_{i} & =\mathbf{R}_{i}^{T}\left(\boldsymbol{\Omega}_{i-1}\mathbf{R}_{i}+\dot{\mathbf{R}}_{i}\right)\\
\boldsymbol{\upsilon}_{i} & =\mathbf{R}_{i}^{T}\left(\boldsymbol{\upsilon}_{i-1}+\boldsymbol{\Omega}_{i-1}\boldsymbol{p}_{i}+\dot{\boldsymbol{p}}_{i}\right)
\end{split}
\label{eq:wvi_recursive}
\end{align}
where we define $\boldsymbol{\Omega}_{i}\left(\boldsymbol{q}^{i},\dot{\boldsymbol{q}}^{i}\right)\in\mathbb{R}^{3\times3}$
and $\boldsymbol{\omega}_{i}=\boldsymbol{\Omega}_{i}^{\vee}$ for
ease of subsequent development of the EoM, The derivations are outlined
in appendices \ref{a:vi_derivation} and \ref{a:Oi_derivation}.%

As shown in \cite{godage2016dynamics}, Jacobians and Hessians play
a critical role in recursive development of the EoM. Applying the
standard techniques, the angular and linear velocity Jacobians, $\mathbf{J}_{i}^{\omega}\left(\boldsymbol{q}^{i},\xi_{i}\right)\in\mathbb{R}^{3\times3n}$
and $\mathbf{J}_{i}^{\upsilon}\left(\boldsymbol{q}^{i},\xi_{i}\right)\in\mathbb{R}^{3\times3n}$
respectively are derived. Here also, we use the property $\boldsymbol{\omega}_{i}=\boldsymbol{\Omega}_{i}^{\vee}$
to define $\mathbf{J}_{i}^{\Omega}\left(\boldsymbol{q}^{i},\xi_{i}\right)\in\mathbb{R}^{3\times9n}$,%
{} in the development of the EoM and it is given by %

\begin{align}
\mathbf{J}_{i}^{\Omega} & =\mathbf{R}_{i}^{T}\left[\begin{array}{c|c}
\mathbf{J}_{i-1}^{\Omega}\mathbf{R}_{i} & \mathbf{R}_{i,\boldsymbol{q}_{i}}\end{array}\right]\label{eq:Jo}
\end{align}
where %
and $\mathbf{J}_{i}^{\omega}=\left(\mathbf{J}_{i}^{\Omega}\right)^{\vee}$
and $\mathbf{J}_{i-1}^{\Omega}\left(\boldsymbol{q}^{i-1}\right)\in\mathbb{R}^{3\times9\left(n-1\right)}$.
Appendix \ref{a:Joi_derivation} details the derivation.

Taking the partial derivative of \eqref{eq:Jo} with respect to $\boldsymbol{q}^{i}$,
the angular body velocity Hessian,%
{} $\mathbf{H}_{i}^{\Omega}=\mathbf{J}_{i,\boldsymbol{q}_{i}}^{\Omega}\left(\boldsymbol{q}^{i},\xi_{i}\right)\in\mathbb{R}^{9n\times9n}$
is given by

\begin{align}
\mathbf{H}_{i}^{\Omega} & =\left[\begin{array}{c|c}
\mathbf{R}_{i}^{T}\mathbf{H}_{i-1}^{\Omega}\mathbf{R}_{i} & \boldsymbol{0}\\
\hline \mathbf{R}_{i,\boldsymbol{q}_{i}}^{T}\mathbf{J}_{i-1}^{\Omega}\mathbf{R}_{i}\cdots & \mathbf{R}_{i,\boldsymbol{q}_{i}}^{T}\mathbf{R}_{i,\boldsymbol{q}_{i}^{T}}\cdots\\
\quad+\mathbf{R}_{i}^{T}\mathbf{J}_{i-1}^{\Omega}\mathbf{R}_{i,\boldsymbol{q}_{i}} & \quad+\mathbf{R}_{i}^{T}\mathbf{R}_{i,\boldsymbol{q}_{i}^{T},\boldsymbol{q}_{i}}
\end{array}\right]\label{eq:Ho}
\end{align}
where $\mathbf{H}_{i-1}^{\Omega}\left(\boldsymbol{q}^{i-1}\right)\in\mathbb{R}^{9\left(n-1\right)\times9\left(n-1\right)}$.
Refer to Appendix \ref{a:Hoi_derivation} for the derivation.%
{} 

Similarly, the linear velocity Jacobian, $\mathbf{J}_{i}^{\upsilon}$,
and Hessian, $\mathbf{H}_{i}^{\upsilon}=\mathbf{J}_{i,\boldsymbol{q}_{i}}^{\upsilon}\left(\boldsymbol{q}^{i},\xi_{i}\right)\in\mathbb{R}^{9n\times3n}$
are given by 

\begin{align}
\mathbf{J}_{i}^{\upsilon} & =\mathbf{R}_{i}^{T}\left[\begin{array}{c|c}
\mathbf{J}_{i-1}^{\upsilon}+\mathbf{J}_{i-1}^{\Omega}\boldsymbol{p}_{i} & \boldsymbol{p}_{i,\boldsymbol{q}_{i}^{T}}\end{array}\right]\label{eq:Jv}
\end{align}

\begin{align}
\mathbf{H}_{i}^{\upsilon} & =\left[\begin{array}{c|c}
\mathbf{R}_{i}^{T}\left(\mathbf{H}_{i-1}^{\upsilon}+\mathbf{H}_{i-1}^{\Omega}\boldsymbol{p}_{i}\right) & \boldsymbol{0}\\
\hline \mathbf{R}_{i,\boldsymbol{q}_{i}}^{T}\left(\mathbf{J}_{i-1}^{\upsilon}+\mathbf{J}_{i-1}^{\Omega}\boldsymbol{p}_{i}\right)\cdots & \mathbf{R}_{i,\boldsymbol{q}_{i}}^{T}\boldsymbol{p}_{i,\boldsymbol{q}_{i}^{T}}\cdots\\
\quad+\mathbf{R}_{i}^{T}\mathbf{J}_{i-1}^{\Omega}\boldsymbol{p}_{i,\boldsymbol{q}_{i}} & \quad+\mathbf{R}_{i}^{T}\boldsymbol{p}_{i,\boldsymbol{q}_{i}^{T},\boldsymbol{q}_{i}}
\end{array}\right]\label{eq:Hv}
\end{align}
where $\mathbf{J}_{i-1}^{\upsilon}\left(\boldsymbol{q}^{i-1},\xi_{i}\right)\in\mathbb{R}^{3\times3\left(n-1\right)}$,
$\mathbf{H}_{i-1}^{\upsilon}\left(\boldsymbol{q}^{i-1}\right)\in\mathbb{R}^{9\left(n-1\right)\times3\left(n-1\right)}$
and the derivation is listed in appendices \ref{a:Jvi_derivation}
and \ref{a:Hvi_derivation}.

\subsection{Extension for Kinematics of Centers of Gravity\label{subsec:Extension-for-Kinematics}}

Similar to Section \ref{subsec:Recursive-Velocities,-Jacobians,},
without losing generality, we derive the kinematics for the CoG of
any $i^{th}$ section . %
We define a coordinate system at the CoG, $\left\{ \overline{O}_{i}\right\} $,
whose %
{} HTM, $\overline{\mathbf{T}}_{i}:\left(\boldsymbol{q}_{i}\right)\mapsto\mathbb{SE}^{3}$,
with respect to $\left\{ O_{i}\right\} $ is defined as

\begin{align}
\overline{\mathbf{T}}_{i} & =\int\mathbf{T}_{i}=\left[\begin{array}{cc}
\overline{\mathbf{R}}_{i} & \overline{\boldsymbol{p}}_{i}\\
\boldsymbol{0} & 1
\end{array}\right]\label{eq:HTM_c_section}
\end{align}
where $\overline{\mathbf{R}}_{i}=\int\mathbf{R}_{i}\left(\boldsymbol{q}_{i}\right)\in\mathbb{R}^{3\times3}$
is the resultant rotation matrix and $\overline{\boldsymbol{p}}_{i}=\int\boldsymbol{p}_{i}\left(\boldsymbol{q}_{i}\right)\in\mathbb{R}^{3}$
is the position vector \cite{godage2015efficient}. Note that the
CoG is a function of $\boldsymbol{q}_{i}$ and therefore varies as
the continuum section deforms. 

To derive the kinematics of the CoG coordinate frame, $\left\{ \overline{O}_{i}\right\} $,
with respect to $\left\{ O\right\} $, we can combine $\overline{\mathbf{T}}_{i}$
with the general HTM given in \eqref{eq:HTM_ith}. From the definition,
$\left\{ O_{i-1}'|_{\xi_{i-1}=1}\right\} \equiv\left\{ O_{i}\right\} $
(Fig. \ref{fig:section_slice_vels_forces}) %
and therefore, CoG of the $i^{th}$section relative to $\left\{ O\right\} $,
$\overline{\mathbf{T}}^{i}:\left(\boldsymbol{q}^{i}\right)\mapsto\mathbb{SE}^{3}$,
can be defined as

\begin{align}
\overline{\mathbf{T}}^{i} & =\int\mathbf{T}^{i-1}\mathbf{T}_{i}=\left(\prod_{k=1}^{i-1}\mathbf{T}_{k}\right)\left(\int\mathbf{T}_{i}\right)=\left[\begin{array}{cc}
\overline{\mathbf{R}}^{i} & \overline{\boldsymbol{p}}^{i}\\
\boldsymbol{0} & 1
\end{array}\right]\label{eq:HTM_c_ith}
\end{align}
where $\overline{\mathbf{R}}^{i}\left(\boldsymbol{q}^{i}\right)\in\mathbb{R}^{3\times3}$
is orientation and $\overline{\boldsymbol{p}}_{i}\left(\boldsymbol{q}^{i}\right)\in\mathbb{R}^{3}$
are position matrices of the CoG coordinate frame. 

Analogous to \eqref{eq:pR_recursive}, the recursive form of $\overline{\mathbf{R}}^{i}$
and $\overline{\boldsymbol{p}}^{i}$ are given by

\begin{align}
\begin{split}\overline{\mathbf{R}}^{i} & =\mathbf{R}^{i-1}\overline{\mathbf{R}}_{i}\\
\overline{\boldsymbol{p}}^{i} & =\boldsymbol{p}^{i-1}+\mathbf{R}^{i-1}\overline{\boldsymbol{p}}_{i}
\end{split}
\label{eq:pR_c_recursive}
\end{align}
where $\mathbf{R}^{i-1}$ and $\boldsymbol{p}^{i-1}$ are formulated
from \eqref{eq:pR_recursive}.

Similar to \eqref{eq:wvi_recursive}, the angular and linear body
velocities of the CoG (relative to $\left\{ \overline{O}_{i}\right\} $),
$\boldsymbol{\overline{\boldsymbol{\omega}}}_{i}\left(\boldsymbol{q}^{i},\dot{\boldsymbol{q}}^{i}\right)\in\mathbb{R}^{3}$
and $\overline{\boldsymbol{\upsilon}}_{i}\left(\boldsymbol{q}^{i},\dot{\boldsymbol{q}}^{i}\right)\in\mathbb{R}^{3}$
can be derived as 
\begin{align}
\begin{split}\overline{\boldsymbol{\Omega}}_{i} & =\overline{\mathbf{R}}_{i}^{T}\left(\boldsymbol{\Omega}_{i-1}\overline{\mathbf{R}}_{i}+\dot{\overline{\mathbf{R}}}_{i}\right)\\
\overline{\boldsymbol{\upsilon}}_{i} & =\overline{\mathbf{R}}_{i}^{T}\left(\boldsymbol{\upsilon}_{i-1}+\boldsymbol{\Omega}_{i-1}\overline{\boldsymbol{p}}_{i}+\dot{\boldsymbol{\overline{p}}}_{i}\right)
\end{split}
\label{eq:wvi_c_recursive}
\end{align}
where $\boldsymbol{\upsilon}_{i-1}$ and $\boldsymbol{\Omega}_{i-1}$,
defined in \eqref{eq:wvi_recursive}, are linear and angular velocities
at the tip of the $\left(i-1\right)^{th}$ continuum section. Here
too, we employ the relationship $\overline{\boldsymbol{\omega}}_{i}=\overline{\boldsymbol{\Omega}}_{i}^{\vee}$.
to compute $\overline{\boldsymbol{\Omega}}_{i}\left(\boldsymbol{q}^{i},\dot{\boldsymbol{q}}^{i}\right)\in\mathbb{R}^{3\times3}$.

Akin to \eqref{eq:Jo}, \eqref{eq:Ho}, \eqref{eq:Jv}, and \eqref{eq:Hv}
the angular body velocity Jacobian of CoG, $\overline{\mathbf{J}}_{i}^{\Omega}\left(\boldsymbol{q}^{i}\right)\in\mathbb{R}^{3\times9n}$,
Hessian $\overline{\mathbf{H}}_{i}^{\Omega}\left(\boldsymbol{q}^{i}\right)\in\mathbb{R}^{9n\times9n}$,
linear body velocity Jacobian , $\overline{\mathbf{J}}_{i}^{\upsilon}\left(\boldsymbol{q}^{i}\right)\in\mathbb{R}^{3\times3n}$,
Hessian $\overline{\mathbf{H}}_{i}^{\upsilon}\left(\boldsymbol{q}^{i}\right)\in\mathbb{R}^{9n\times3n}$,
are respectively given by \eqref{eq:Jo_c}, \eqref{eq:Ho_c}, \eqref{eq:Jv_c},
and \eqref{eq:Hv_c} as 

\begin{align}
\overline{\mathbf{J}}_{i}^{\Omega} & =\overline{\mathbf{R}}_{i}^{T}\left[\begin{array}{c|c}
\mathbf{J}_{i-1}^{\Omega}\overline{\mathbf{R}}_{i} & \overline{\mathbf{R}}_{i,\boldsymbol{q}_{i}^{T}}\end{array}\right]\label{eq:Jo_c}\\
\overline{\mathbf{H}}_{i}^{\Omega} & =\left[\begin{array}{c|c}
\overline{\mathbf{R}}_{i}^{T}\mathbf{H}_{i-1}^{\Omega}\overline{\mathbf{R}}_{i} & \boldsymbol{0}\\
\hline \overline{\mathbf{R}}_{i,\boldsymbol{q}_{i}}^{T}\mathbf{J}_{i-1}^{\Omega}\overline{\mathbf{R}}_{i}\cdots & \overline{\mathbf{R}}_{i,\boldsymbol{q}_{i}}^{T}\overline{\mathbf{R}}_{i,\boldsymbol{q}_{i}^{T}}\cdots\\
\;+\overline{\mathbf{R}}_{i}^{T}\mathbf{J}_{i-1}^{\Omega}\overline{\mathbf{R}}_{i,\boldsymbol{q}_{i}} & +\overline{\mathbf{R}}_{i}^{T}\overline{\mathbf{R}}_{i,\boldsymbol{q}_{i}^{T},\boldsymbol{q}_{i}}
\end{array}\right]\label{eq:Ho_c}\\
\overline{\mathbf{J}}_{i}^{\upsilon} & =\overline{\mathbf{R}}_{i}^{T}\left[\begin{array}{c|c}
\mathbf{J}_{i-1}^{\upsilon}+\mathbf{J}_{i-1}^{\Omega}\overline{\boldsymbol{p}}_{i} & \overline{\boldsymbol{p}}_{i,\boldsymbol{q}_{i}^{T}}\end{array}\right]\label{eq:Jv_c}\\
\overline{\mathbf{H}}_{i}^{\upsilon} & =\left[\begin{array}{c|c}
\overline{\mathbf{R}}_{i}^{T}\left(\mathbf{H}_{i-1}^{\upsilon}+\mathbf{H}_{i-1}^{\Omega}\overline{\boldsymbol{p}}_{i}\right) & \boldsymbol{0}\\
\hline \overline{\mathbf{R}}_{i,\boldsymbol{q}_{i}}^{T}\left(\mathbf{J}_{i-1}^{\upsilon}+\mathbf{J}_{i-1}^{\Omega}\overline{\boldsymbol{p}}_{i}\right)\cdots & \overline{\mathbf{R}}_{i,\boldsymbol{q}_{i}}^{T}\overline{\boldsymbol{p}}_{i,\boldsymbol{q}_{i}^{T}}\cdots\\
\;+\overline{\mathbf{R}}_{i}^{T}\mathbf{J}_{i-1}^{\Omega}\overline{\boldsymbol{p}}_{i,\boldsymbol{q}_{i}} & +\overline{\mathbf{R}}_{i}^{T}\overline{\boldsymbol{p}}_{i,\boldsymbol{q}_{i}^{T},\boldsymbol{q}_{i}}
\end{array}\right]\label{eq:Hv_c}
\end{align}

\subsection{Case Study: Point vs. Non-point Mass at the CoG\label{subsec:Case-Study:-Point}}

Consider the CoG velocities depicted in \eqref{eq:wvi_c_recursive}
when $\boldsymbol{\Omega}_{i-1}=\left[0,0,\omega_{z}\right]$ with
$\omega_{z}\ne0$ , $\boldsymbol{\upsilon}_{i-1}=0$, $\boldsymbol{q}_{i}=0$,
and $\dot{\boldsymbol{q}}_{i}=0$. Physically this refers to a non-actuating
$i^{th}$ continuum section (essentially a cylinder of length $L_{i0}$
and radius $r_{i}$ whose CoG is located at the mid point, i.e., $\overline{\boldsymbol{p}}_{i}=\left[0,0,\frac{L_{i0}}{2}\right]$,
of the neutral axis where the tip of the $\left(i-1\right)^{th}$
section rotates in place without translation. This scenario is theoretically
possible and demonstrated in \cite{godage2015modal} where kinematic
decoupling is present in multisection continuum arms. From \eqref{eq:wvi_c_recursive},
the CoG velocities become $\overline{\boldsymbol{\Omega}}_{i}=\boldsymbol{\Omega}_{i-1}$
and $\overline{\boldsymbol{\upsilon}}_{i}=0$. The kinetic energies
of the $i^{th}$ section then become $\mathcal{K}_{i}^{\omega}=\frac{1}{4}m_{i}r_{i}^{2}\omega_{z}^{2}$
and $\mathcal{K}_{i}^{\upsilon}=0$. If a point-mass is considered
at the CoG, it will result in $\mathcal{\overline{K}}_{i}^{\omega}=\overline{\mathcal{K}}_{i}^{\upsilon}=0$.
As a result, it becomes evident that a point-mass model is not suitable
for modeling multisection continuum arms. Thus, in this paper, we
will consider a hypothetical thin disc of mass $m_{i}$ and radius
$r_{i}$ on the XY plane of $\left\{ \overline{O}_{i}\right\} $ with
its geometric center coinciding the origin of $\left\{ \overline{O}_{i}\right\} $,
i.e., at the CoG (Fig. \ref{fig:section_slice_vels_forces}). The
respective kinetic energies then become $\mathcal{\overline{K}}_{i}^{\omega}=\frac{1}{4}m_{i}r_{i}^{2}\omega_{z}^{2}$
and $\overline{\mathcal{K}}_{i}^{\upsilon}=0$ to match that of the
actual continuum section energy. Employing the disc model at the CoG,
following section derives the energy shaping coefficients \cite{godage2015efficient}
to match energies to that of the integral model reported in \cite{godage2016dynamics}.%

\section{Derive Energy Balance of Center of Gravity-based System\label{sec:Balance-Energies-of}}

\subsection{Continuum Section Kinetic Energy: Integral and CoG-based Models \label{subsec:Energies-of-Continuum}}

Without losing generality, we next derive the kinetic energies, angular
and linear, for any $i^{th}$ continuum section. Then we compare the
terms to formulate the energy scaling conditions. Analogous to \cite{godage2016dynamics},
to find the kinetic energy of the continuum section using an integral
approach, we will consider an infinitesimally thin disc of radius
$r_{i}$ along the length of the continuum section. By applying the
body velocities given by equations \eqref{eq:wvi_recursive}, the
energy computed for a disc is then integrated with respect to $\xi_{i}$
to compute the section energy%
. The angular kinetic energy, $\mathcal{K}_{i}^{\omega}:\left(\boldsymbol{q}^{i},\dot{\boldsymbol{q}}^{i}\right)\mapsto\mathbb{R}$,
is given by

\begin{align}
\mathcal{K}_{i}^{\omega} & =\int\left(\frac{1}{2}\boldsymbol{\omega}_{i}^{T}\mathcal{M}_{i}^{\omega}\boldsymbol{\omega}_{i}\right)=\frac{1}{2}I_{xx}\mathbb{T}_{2}\left(\int\boldsymbol{\Omega}_{i}^{T}\boldsymbol{\Omega}_{i}\right)\nonumber \\
 & =\frac{1}{2}I_{xx}\mathbb{T}_{2}\left(\int\mathbf{R}_{i}^{T}\boldsymbol{\Omega}_{i-1}^{T}\boldsymbol{\Omega}_{i-1}\mathbf{R}_{i}\right.\cdots\nonumber \\
 & \qquad\left.+2\int\dot{\mathbf{R}}_{i}^{T}\boldsymbol{\Omega}_{i-1}\mathbf{R}_{i}+\int\dot{\mathbf{R}}_{i}^{T}\dot{\mathbf{R}}_{i}\right)\label{eq:Kwi}
\end{align}
where $I_{xx}=\frac{1}{4}m_{i}r_{i}^{2}$ is the moment of inertia
about the X axis of $\left\{ O_{i}'\right\} $.

Using the angular velocity given in \eqref{eq:wvi_c_recursive}, finding
the angular kinetic energy of the disc at the CoG, $\overline{\mathcal{K}}_{i}^{\omega}:\left(\boldsymbol{q}^{i},\dot{\boldsymbol{q}}^{i}\right)\mapsto\mathbb{R}_{0}^{+}$,
results in

\begin{align}
\overline{\mathcal{K}}_{i}^{\omega} & =\frac{1}{2}\overline{\boldsymbol{\omega}}_{i}^{T}\mathcal{M}_{i}^{\omega}\overline{\boldsymbol{\omega}}_{i}=\frac{1}{2}I_{xx}\mathbb{T}_{2}\left(\overline{\boldsymbol{\Omega}}_{i}^{T}\overline{\boldsymbol{\Omega}}_{i}\right)\label{eq:Kwi_c}\\
 & =\frac{1}{2}I_{xx}\mathbb{T}_{2}\left(\overline{\mathbf{R}}_{i}^{T}\boldsymbol{\Omega}_{i-1}^{T}\boldsymbol{\Omega}_{i-1}\overline{\mathbf{R}}_{i}+2\dot{\overline{\mathbf{R}}}_{i}^{T}\boldsymbol{\Omega}_{i-1}\overline{\mathbf{R}}_{i}+\dot{\overline{\mathbf{R}}}_{i}^{T}\dot{\overline{\mathbf{R}}}_{i}\right)\nonumber 
\end{align}

Similarly, using the linear body velocity in \eqref{eq:wvi_recursive},
the linear kinetic energy of the continuous model, $\mathcal{K}_{i}^{\upsilon}:\left(\boldsymbol{q}^{i},\dot{\boldsymbol{q}}^{i}\right)\mapsto\mathbb{R}_{0}^{+}$,
can be computed as

\begin{align}
\mathcal{K}_{i}^{\upsilon} & =\int\left(\frac{1}{2}\boldsymbol{\upsilon}_{i}^{T}\mathcal{M}_{i}^{\upsilon}\boldsymbol{\upsilon}_{i}\right)\label{eq:Kvi}\\
 & =\frac{1}{2}m_{i}\left(\boldsymbol{\upsilon}_{i-1}^{T}\boldsymbol{\upsilon}_{i-1}+2\boldsymbol{\upsilon}_{i-1}^{T}\boldsymbol{\Omega}_{i-1}\overline{\boldsymbol{p}}_{i}+2\boldsymbol{\upsilon}_{i-1}^{T}\dot{\overline{\boldsymbol{p}}}_{i}\cdots\right.\nonumber \\
 & \;\left.+\int\boldsymbol{p}_{i}^{T}\boldsymbol{\Omega}_{i-1}^{T}\boldsymbol{\Omega}_{i-1}\boldsymbol{p}_{i}+2\int\boldsymbol{p}_{i}^{T}\boldsymbol{\Omega}_{i-1}^{T}\dot{\boldsymbol{p}}_{i}+\int\dot{\boldsymbol{p}}_{i}^{T}\dot{\boldsymbol{p}}_{i}\right)\nonumber 
\end{align}
where $\mathcal{M}_{i}^{\upsilon}=m_{i}\mathbf{I}_{3}$. Additionally,
the CoG model's linear kinetic energy, $\overline{\mathcal{K}}_{i}^{\upsilon}:\left(\boldsymbol{q}^{i},\dot{\boldsymbol{q}}^{i}\right)\mapsto\mathbb{R}_{0}^{+}$,
is derived as

\begin{align}
\overline{\mathcal{K}}_{i}^{\upsilon} & =\frac{1}{2}\overline{\boldsymbol{\upsilon}}_{i}^{T}\mathcal{M}_{i}^{\upsilon}\overline{\boldsymbol{\upsilon}}_{i}=\frac{1}{2}m_{i}\left(\boldsymbol{\upsilon}_{i-1}^{T}\boldsymbol{\upsilon}_{i-1}+2\boldsymbol{\upsilon}_{i-1}^{T}\boldsymbol{\Omega}_{i-1}\overline{\boldsymbol{p}}_{i}\right.\cdots\nonumber \\
 & \;\left.+2\boldsymbol{\upsilon}_{i-1}^{T}\dot{\overline{\boldsymbol{p}}}_{i}+\overline{\boldsymbol{p}}_{i}^{T}\boldsymbol{\Omega}_{i-1}^{T}\boldsymbol{\Omega}_{i-1}\overline{\boldsymbol{p}}_{i}+2\overline{\boldsymbol{p}}_{i}^{T}\boldsymbol{\Omega}_{i-1}^{T}\dot{\overline{\boldsymbol{p}}}_{i}+\dot{\overline{\boldsymbol{p}}}_{i}^{T}\dot{\overline{\boldsymbol{p}}}_{i}\right)\label{eq:Kvi_c}
\end{align}

\subsection{Minimize Energy Difference Between the Integral and CoG-based Models\label{subsec:Minimize-Energy-Difference}}

In this section, %
utilizing the energies derived in Section \ref{subsec:Energies-of-Continuum},
we systematically derive scalars to match the kinetic energy of the
CoG models to that of the integral model. Unlike the single section
case \cite{godage2015efficient} however, the kinetic energy is dependent
on the velocities of the $i^{th}$ section as well as the previous
sections. Consider the angular energy difference between the models,
derived for the $i^{th}$ continuum section%
, given by

\begin{align}
\mathcal{K}_{i}^{\omega}-\overline{\mathcal{K}}_{i}^{\omega} & =\frac{1}{2}I_{xx}\mathbb{T}_{2}\left(\int\dot{\mathbf{R}}_{i}^{T}\dot{\mathbf{R}}_{i}-\beta_{3}^{\omega}\dot{\overline{\mathbf{R}}}_{i}^{T}\dot{\overline{\mathbf{R}}}_{i}\cdots\right.\nonumber \\
 & +2\int\!\!\mathbf{R}_{i}^{T}\boldsymbol{\Omega}_{i-1}^{T}\boldsymbol{\Omega}_{i-1}\mathbf{R}_{i}-2\beta_{1}^{\omega}\overline{\mathbf{R}}_{i}^{T}\boldsymbol{\Omega}_{i-1}^{T}\boldsymbol{\Omega}_{i-1}\overline{\mathbf{R}}_{i}\cdots\nonumber \\
 & \quad\left.+\int\dot{\mathbf{R}}_{i}^{T}\boldsymbol{\Omega}_{i-1}\mathbf{R}_{i}-\beta_{2}^{\omega}\dot{\overline{\mathbf{R}}}_{i}^{T}\boldsymbol{\Omega}_{i-1}\overline{\mathbf{R}}_{i}\right)\label{eq:dKw_i}
\end{align}
where $\beta_{k}^{\omega}\,\forall k\in\left\{ 1,2,3\right\} $ are
the energy shaping coefficients that we apply to the CoG energy terms
to match the energies. 

Note that, in this case, unlike the single section case \cite{godage2015accurate},
we have three terms that do not get canceled when taking the difference.
Likewise, the linear kinetic energy difference is computed as

\begin{align}
\mathcal{K}_{i}^{\upsilon}-\overline{\mathcal{K}}_{i}^{\upsilon} & =\frac{1}{2}m_{i}\left(\int\boldsymbol{p}_{i}^{T}\boldsymbol{\Omega}_{i-1}^{T}\boldsymbol{\Omega}_{i-1}\boldsymbol{p}_{i}-\beta_{1}^{\upsilon}\overline{\boldsymbol{p}}_{i}^{T}\boldsymbol{\Omega}_{i-1}^{T}\boldsymbol{\Omega}_{i-1}\overline{\boldsymbol{p}}_{i}\cdots\right.\nonumber \\
 & \quad+\int\boldsymbol{p}_{i}^{T}\boldsymbol{\Omega}_{i-1}^{T}\dot{\boldsymbol{p}}_{i}-\beta_{2}^{\upsilon}\overline{\boldsymbol{p}}_{i}^{T}\boldsymbol{\Omega}_{i-1}^{T}\dot{\overline{\boldsymbol{p}}}_{i}\cdots\nonumber \\
 & \qquad+\left.\int\dot{\boldsymbol{p}}_{i}^{T}\dot{\boldsymbol{p}}_{i}-\beta_{3}^{\upsilon}\dot{\overline{\boldsymbol{p}}}_{i}^{T}\dot{\overline{\boldsymbol{p}}}_{i}\right)\label{eq:dKv_i}
\end{align}

Notice that some terms are canceled due to the absence of products
of integrable terms, and thus resulting in three remaining terms.
We introduce the energy shaping coefficients, $\beta_{k}^{\upsilon}\,\forall k\in\left\{ 1,2,3\right\} $,
for each of those terms. %
The coefficients, introduced in \eqref{eq:dKw_i} and \eqref{eq:dKv_i},
will be solved in the latter part of this section through a multivariate
optimization routine.%

\subsubsection{Generate Random Sample Set\label{subsec:Generate-Random-Sample}}

Including the physical robot parameters such as $L_{i0}$, $l_{i}$,
and $r_{i}$, the energy differences given by \eqref{eq:dKw_i} and
\eqref{eq:dKv_i}, become functions of $\left(\alpha_{l},\alpha_{r},\boldsymbol{q}_{i},\dot{\boldsymbol{q}}_{i},\boldsymbol{\Omega}_{i-1}\right)\in\mathbb{R}^{11}$
where $\alpha_{l}=\frac{\text{max}\left(l_{i}\right)}{L_{i0}}$ and
$\alpha_{r}=\frac{r_{i}}{L_{i0}}$ are the normalized length and radius
of the continuum section. %
Similarly, we generate $10^{6}$ random combinations of $\alpha_{r}\in{\scriptstyle \left[\frac{1}{20},\frac{1}{2}\right]}$,
$\alpha_{l}\in{\scriptstyle \left[\frac{1}{20},6\pi\alpha_{r}\right]}$,
$\boldsymbol{q}_{i}\in\left[0,\alpha_{l}L_{i0}\right]$, and $\dot{\boldsymbol{q}}_{i}\in\left[0,L_{i0}\right]$.
The upper bound of $\alpha_{l}$ limits maximum bending angle of continuum
sections to $\frac{4\pi}{3}$. Also, note that $\boldsymbol{\Omega}_{i-1}$
depends on $\left(\boldsymbol{q}^{i-1},\dot{\boldsymbol{q}}^{i-1}\right)$,
and for a general $i^{th}$ section, it is not possible to sample
the joint-space variables since $i$ is arbitrary. To overcome this
challenge, we generate random $\boldsymbol{\Omega}_{i-1}$ where each
component is chosen from the range $\left[-10^{2},10^{2}\right]$.
Note that these parametric bounds for $\boldsymbol{\Omega}_{i-1}$
and $\alpha_{l}$, though arbitrary and unrealistically large for
physical continuum arms, were chosen to ensure the rigor and generality
of the proposed model and within the error bounds of the $13^{th}$
order modal shape functions used in this paper. However, %
one may increase this bound (which would also require adjusting the
order of modal shape functions of the %
HTM elements to meet the desired position and orientation error metrics
at the tip at the maximum bending). More details related on choosing
expansion order and errors can be found in \cite{godage2015modal}.%

\subsubsection{Computing the Energy Shaping Coefficients\label{subsec:Finding-the-Scaling}}

\begin{figure}[tb]
\begin{centering}
\subfloat[]{\begin{centering}
\includegraphics[bb=20bp 10bp 400bp 295bp,width=0.48\columnwidth]{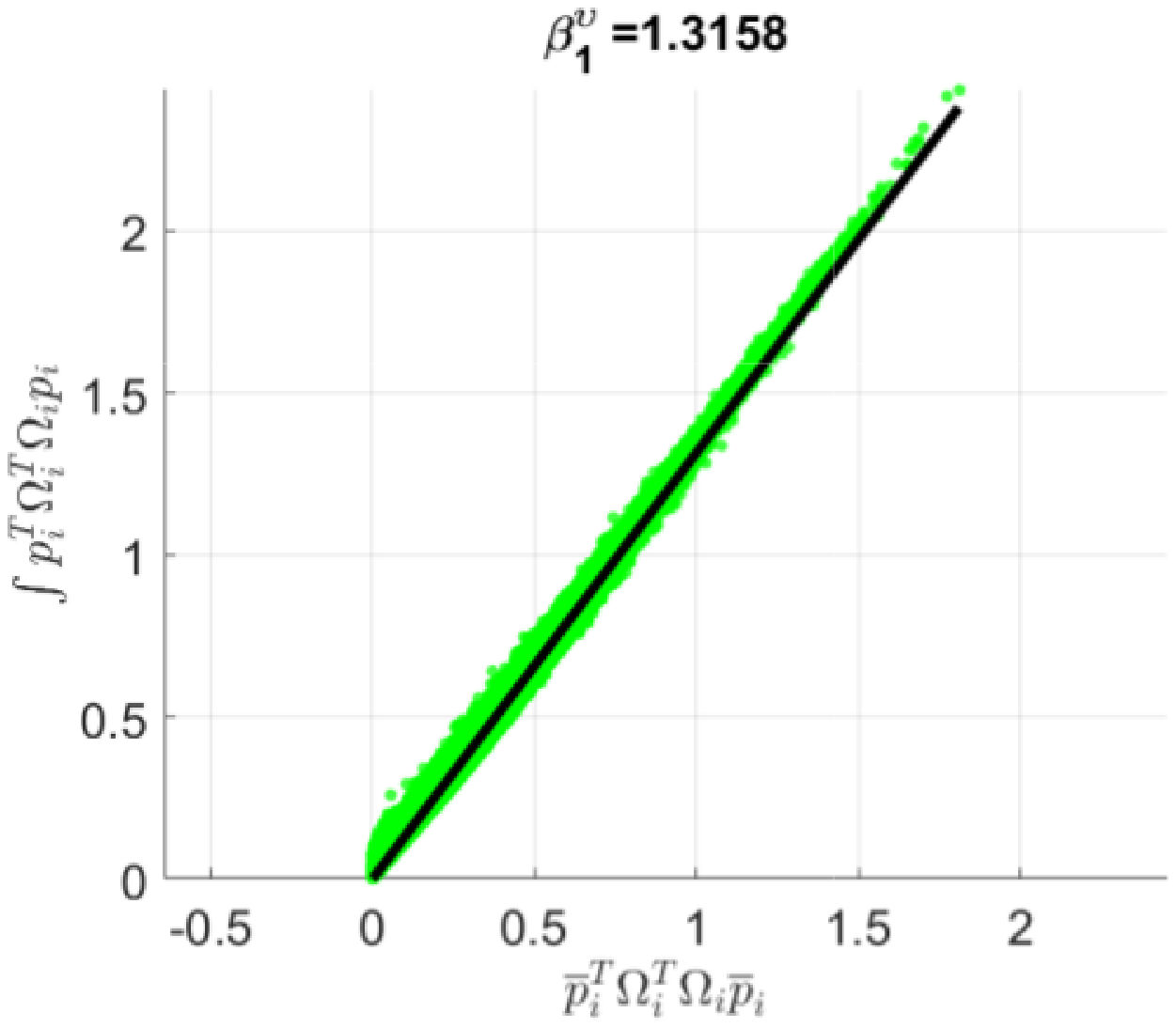}
\par\end{centering}
}\subfloat[]{\begin{centering}
\includegraphics[bb=20bp 10bp 400bp 295bp,width=0.48\columnwidth]{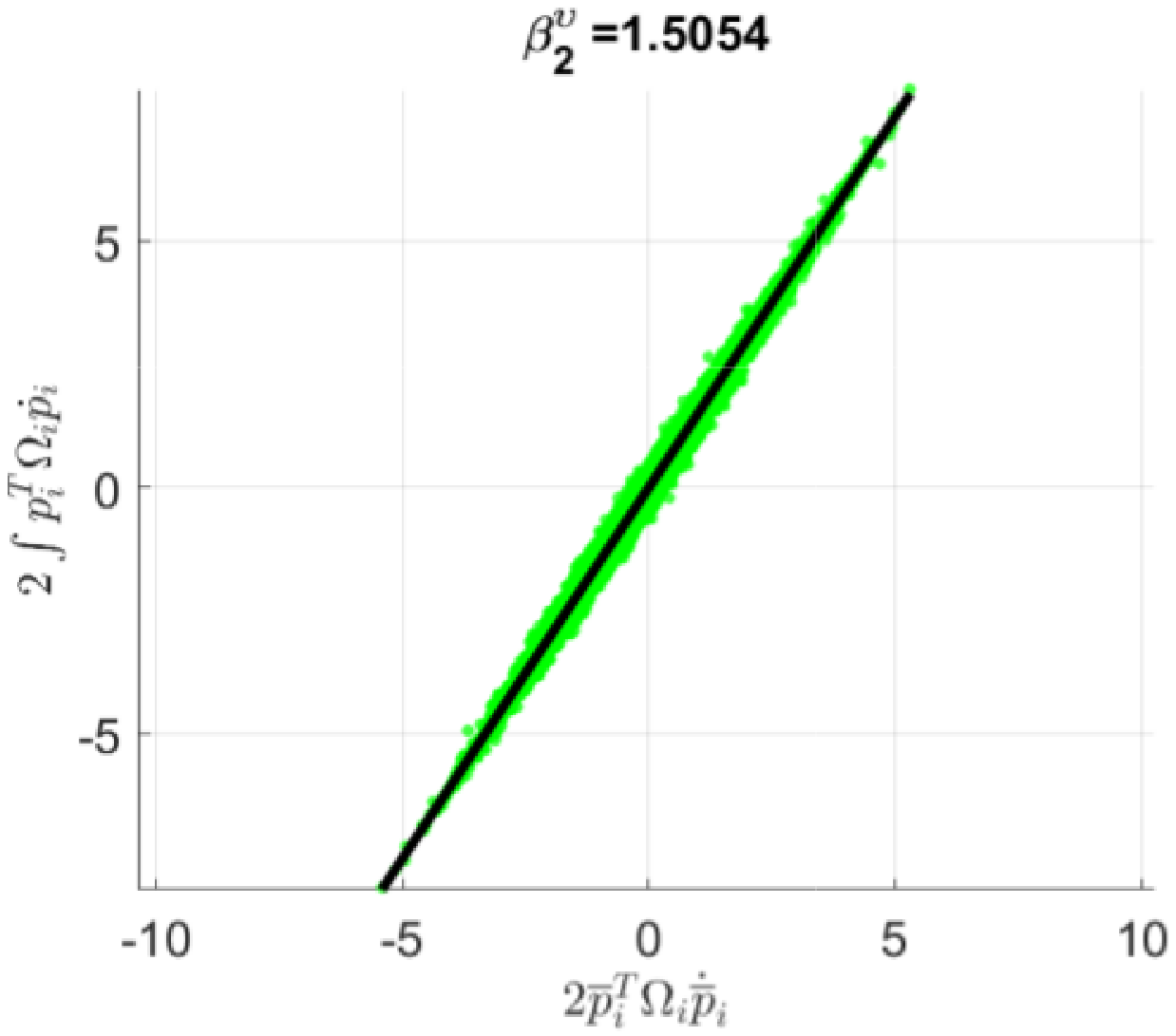}
\par\end{centering}
}
\par\end{centering}
\begin{centering}
\vspace{-0mm}\subfloat[]{\begin{centering}
\includegraphics[bb=20bp 10bp 400bp 295bp,width=0.48\columnwidth]{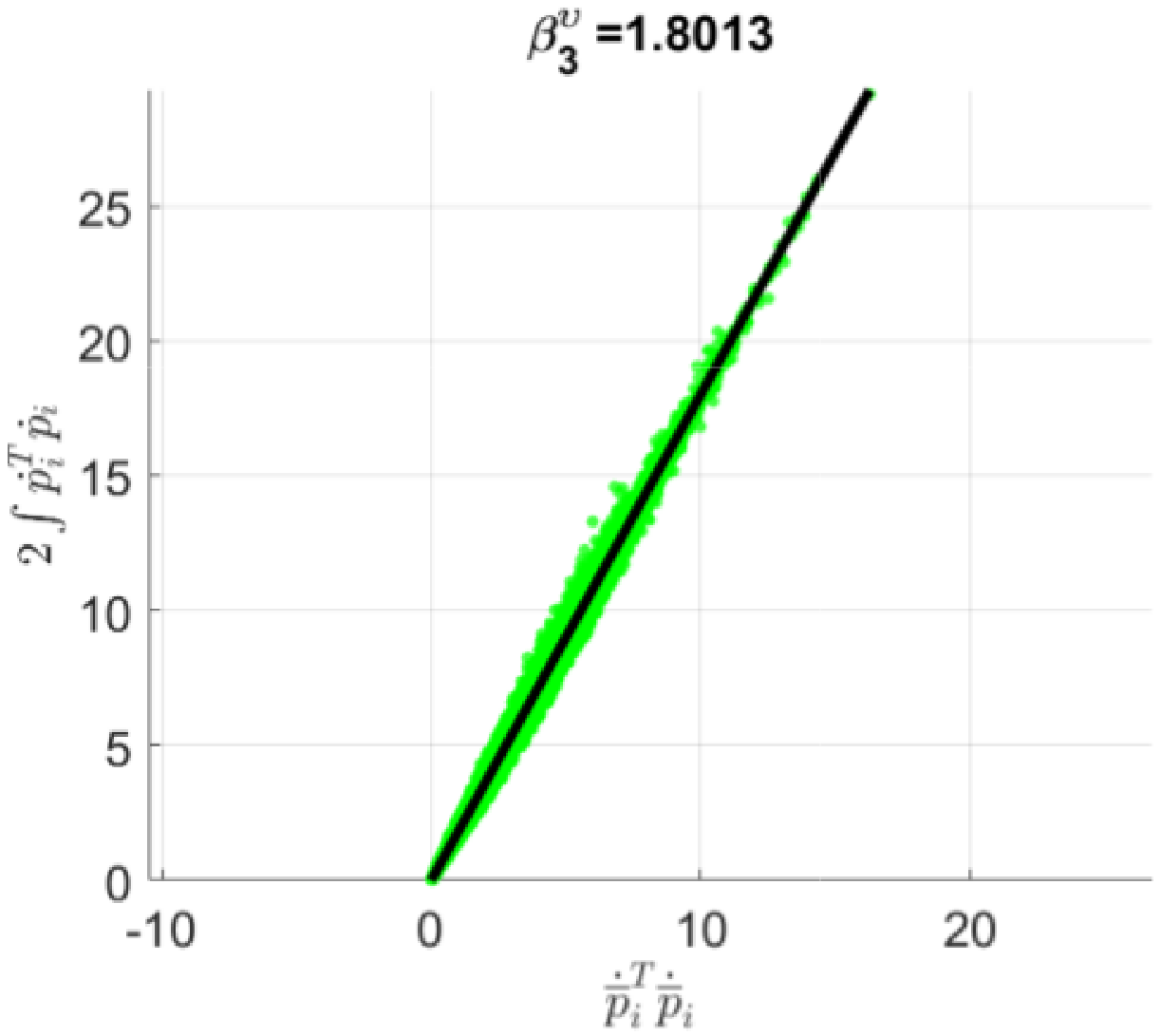}
\par\end{centering}
}\subfloat[]{\begin{centering}
\includegraphics[bb=20bp 10bp 400bp 295bp,width=0.48\columnwidth]{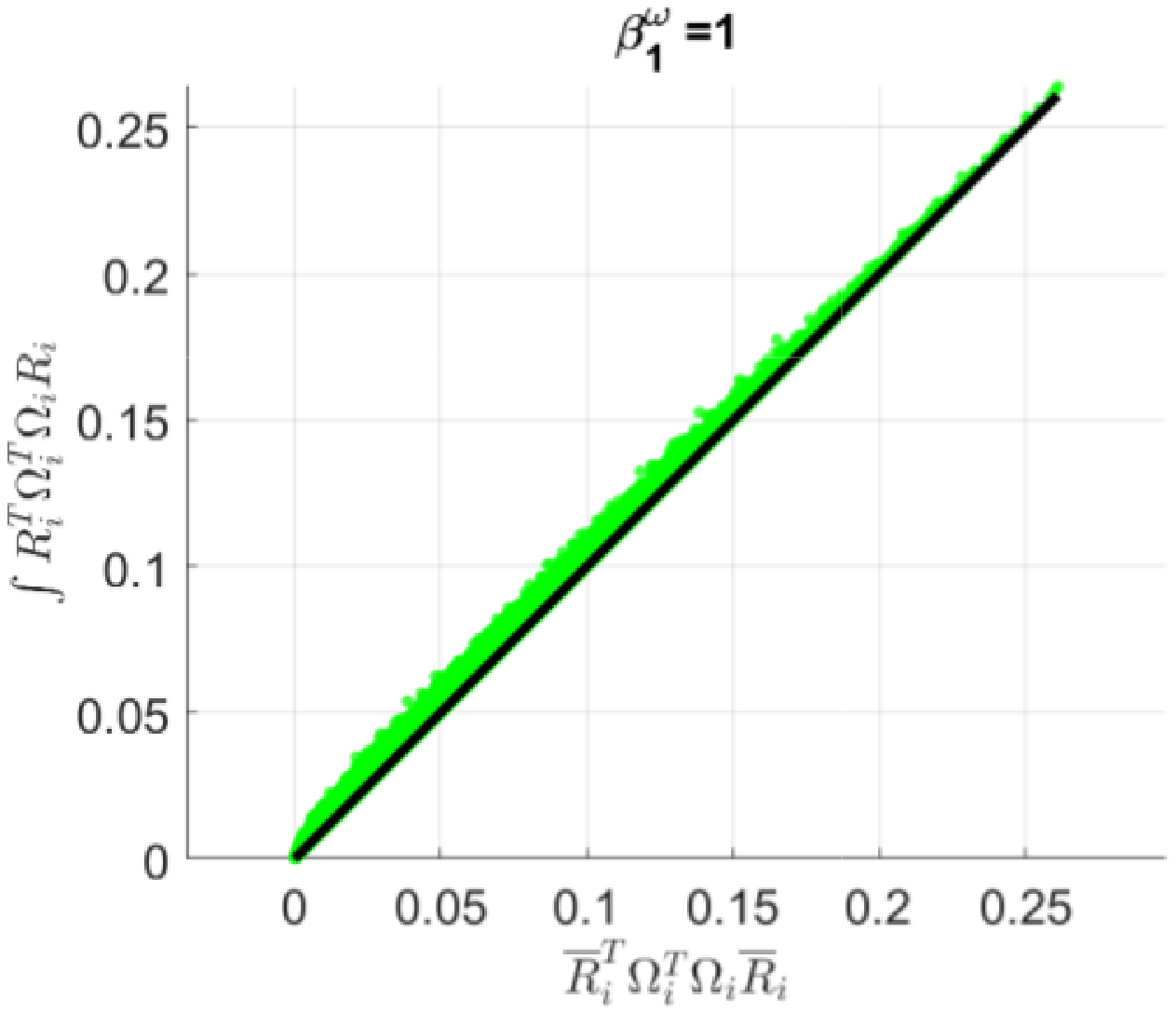}
\par\end{centering}
}
\par\end{centering}
\begin{centering}
\vspace{-0mm}\subfloat[]{\begin{centering}
\includegraphics[bb=20bp 10bp 400bp 295bp,width=0.48\columnwidth]{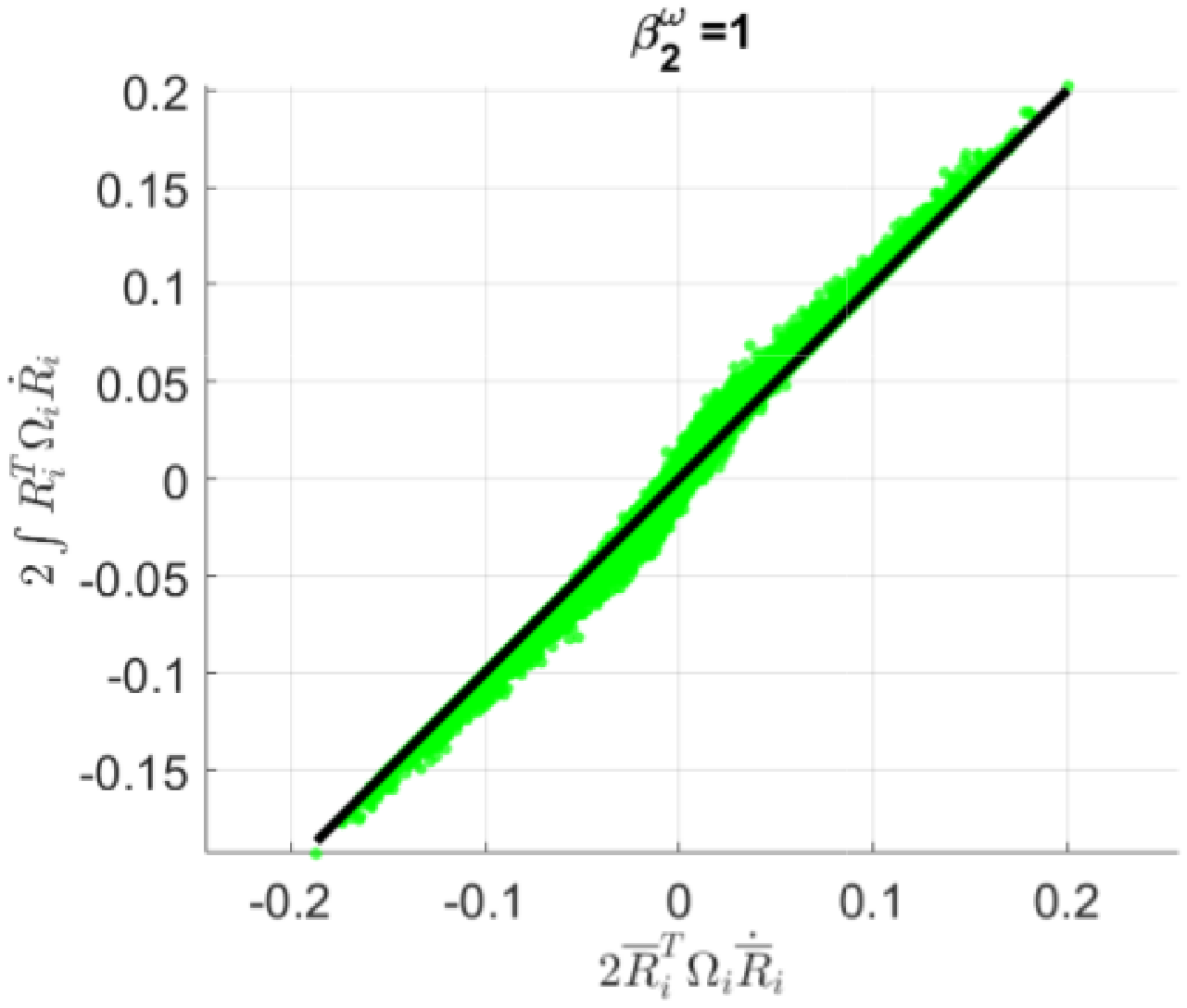}
\par\end{centering}
}\subfloat[]{\begin{centering}
\includegraphics[bb=0bp 10bp 400bp 295bp,width=0.48\columnwidth]{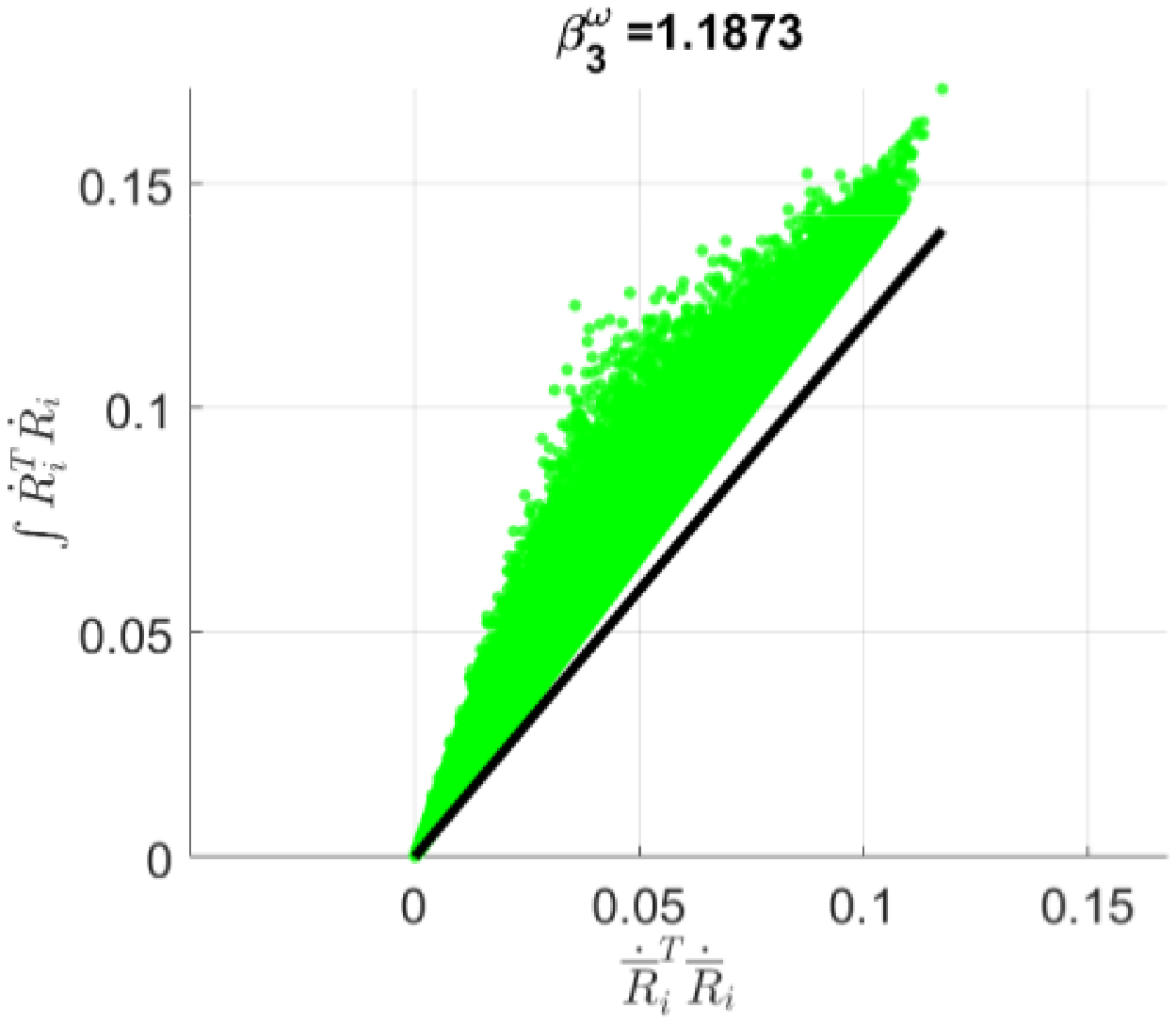}
\par\end{centering}
\label{sfig:bw3}}
\par\end{centering}
\caption{Comparison of the ratio of energy terms given by \eqref{eq:dKv_i}:
(a) $\int\boldsymbol{p}_{i}^{T}\boldsymbol{\Omega}_{i-1}^{T}\boldsymbol{\Omega}_{i-1}\boldsymbol{p}_{i}$
vs. $\overline{\boldsymbol{p}}_{i}^{T}\boldsymbol{\Omega}_{i-1}^{T}\boldsymbol{\Omega}_{i-1}\overline{\boldsymbol{p}}_{i}$,
(b) $\int\boldsymbol{p}_{i}^{T}\boldsymbol{\Omega}_{i-1}^{T}\dot{\boldsymbol{p}}_{i}$
vs. $\overline{\boldsymbol{p}}_{i}^{T}\boldsymbol{\Omega}_{i-1}^{T}\dot{\overline{\boldsymbol{p}}}_{i}$,
(c) $\int\dot{\boldsymbol{p}}_{i}^{T}\dot{\boldsymbol{p}}_{i}$ vs.
$\dot{\overline{\boldsymbol{p}}}_{i}^{T}\dot{\overline{\boldsymbol{p}}}_{i}$.
Similarly, the comparison of the ratio of energy terms given by \eqref{eq:dKw_i}:
(d) $\int\dot{\mathbf{R}}_{i}^{T}\dot{\mathbf{R}}_{i}$ vs. $\dot{\overline{\mathbf{R}}}_{i}^{T}\dot{\overline{\mathbf{R}}}_{i}$,
(e) $2\int\!\!\mathbf{R}_{i}^{T}\boldsymbol{\Omega}_{i-1}^{T}\boldsymbol{\Omega}_{i-1}\mathbf{R}_{i}$
vs. $2\overline{\mathbf{R}}_{i}^{T}\boldsymbol{\Omega}_{i-1}^{T}\boldsymbol{\Omega}_{i-1}\overline{\mathbf{R}}_{i}$,
and (f) $\int\dot{\mathbf{R}}_{i}^{T}\boldsymbol{\Omega}_{i-1}\mathbf{R}_{i}$
vs. ~$\dot{\overline{\mathbf{R}}}_{i}^{T}\boldsymbol{\Omega}_{i-1}\overline{\mathbf{R}}_{i}$. }
\label{fig:Kb_K_energy_comparison}
\end{figure}

For the random combinations of joint-space variables and physical
parameters generated in the previous step, corresponding kinetic energy
differences of the integral and CoG-based models, depicted in \eqref{eq:dKw_i}
and \eqref{eq:dKv_i} are computed. As suggested by the definitions,
for the ease of comparison of corresponding terms, we computed the
three residual terms of each of kinetic energy differences separately.
For instance, in the case of $\mathcal{K}_{i}^{\omega}$, terms ${\scriptstyle {\textstyle \mathbb{T}_{2}(\int\mathbf{R}_{i}^{T}\boldsymbol{\Omega}_{i-1}^{T}\boldsymbol{\Omega}_{i-1}\mathbf{R}_{i})}}$,
$\mathbb{T}_{2}(\int\dot{\mathbf{R}}_{i}^{T}\boldsymbol{\Omega}_{i-1}\mathbf{R}_{i})$,
and $\mathbb{T}_{2}(\int\dot{\mathbf{R}}_{i}^{T}\dot{\mathbf{R}}_{i})$
are computed separately. Similarly, for $\overline{\mathcal{K}}_{i}^{\omega}$,
${\scriptstyle {\textstyle \mathbb{T}_{2}(\overline{\mathbf{R}}_{i}^{T}\boldsymbol{\Omega}_{i-1}^{T}\boldsymbol{\Omega}_{i-1}\overline{\mathbf{R}}_{i})}}$,
$2\mathbb{T}_{2}(\dot{\mathbf{\overline{R}}}_{i}^{T}\boldsymbol{\Omega}_{i-1}\mathbf{\overline{R}}_{i})$,
and $\mathbb{T}_{2}(\dot{\overline{\mathbf{R}}}_{i}^{T}\dot{\mathbf{\overline{R}}}_{i})$
are computed separately. Then the sum of these terms, scaled by $\frac{1}{2}I_{xx}$,
will yield the energy difference, $\mathcal{K}_{i}^{\omega}-\overline{\mathcal{K}}_{i}^{\omega}$.
The same approach is followed for the linear kinetic energy difference
given by \eqref{eq:dKv_i} and scaled by $\frac{m_{i}}{2}$. %
The corresponding terms for the integral system and the CoG-based
system are then plotted against each other in Fig. \ref{fig:Kb_K_energy_comparison}. 

It can be seen that, despite the variation of the physical shape ($\max\left(l_{ij}\right)$
and $r_{i}$), there are proportional relationships between the matching
terms of the two analytical models. This indicates us that the fundamental
variable-length continuum section behavior across the two systems
are proportional and independent of the physical shape. %
The proportional constants can be computed in two ways. One approach
is to consider matching terms individually and compute the least square
linear fit. The other approach is to consider the entire system's
kinetic energy and find the optimal coefficients that would minimize
the cumulative energy difference. In this work, we have opted for
the latter approach, since it provided a slight, though negligible,
improvement in energy matching. We formulated our optimization problem
in Matlab 2017a and used global optimization on the inbuilt fmincon
multivariate constrained optimization subroutine using the objective
function $\mathcal{K}_{i}^{\upsilon}-\overline{\mathcal{K}}_{i}^{\upsilon}\left(\beta^{\upsilon}\right)+\mathcal{K}_{i}^{\omega}-\overline{\mathcal{K}}_{i}^{\omega}\left(\beta^{\omega}\right)$
for all the $10^{6}$ parametric combinations. Noting the direct proportionality,
we bounded the scalar range to $\left[0,1\right]$ for numerical efficiency.
The resultant energy shaping coefficient values are shown in Fig.
\ref{fig:Kb_K_energy_comparison}. 

Notice that the proportional coefficient of Fig. \ref{sfig:bw3} is
slightly more aggressive than what the data suggests. The reason is
the difference of the ratio of contributions from individual terms.
For instance, the contribution of the term $\frac{m_{i}}{2}\int\dot{\boldsymbol{p}}_{i}^{T}\dot{\boldsymbol{p}}_{i}$
is orders of magnitude greater than that of the term $\frac{1}{2}I_{xx}\mathbb{T}_{2}\left(\int\dot{\mathbf{R}}_{i}^{T}\dot{\mathbf{R}}_{i}\right)$.
The system-wide energy consideration would then place more emphasize
on larger contributors to yield optimal energy scalars, and this explains
the sub-optimal results of the term-wise computation of proportional
coefficients. %
However, based on our computations, given the strong correlation of
the energy terms between the two modeling approaches, either method
produces sufficient accuracy for practical purposes.

\subsubsection{Numerical Validation of Energy Shaping Coefficients\label{subsec:Numerical-Validation-of}}

\begin{figure}[tb]
\begin{centering}
\subfloat[]{\begin{centering}
\includegraphics[bb=0bp 0bp 400bp 315bp,width=0.48\columnwidth]{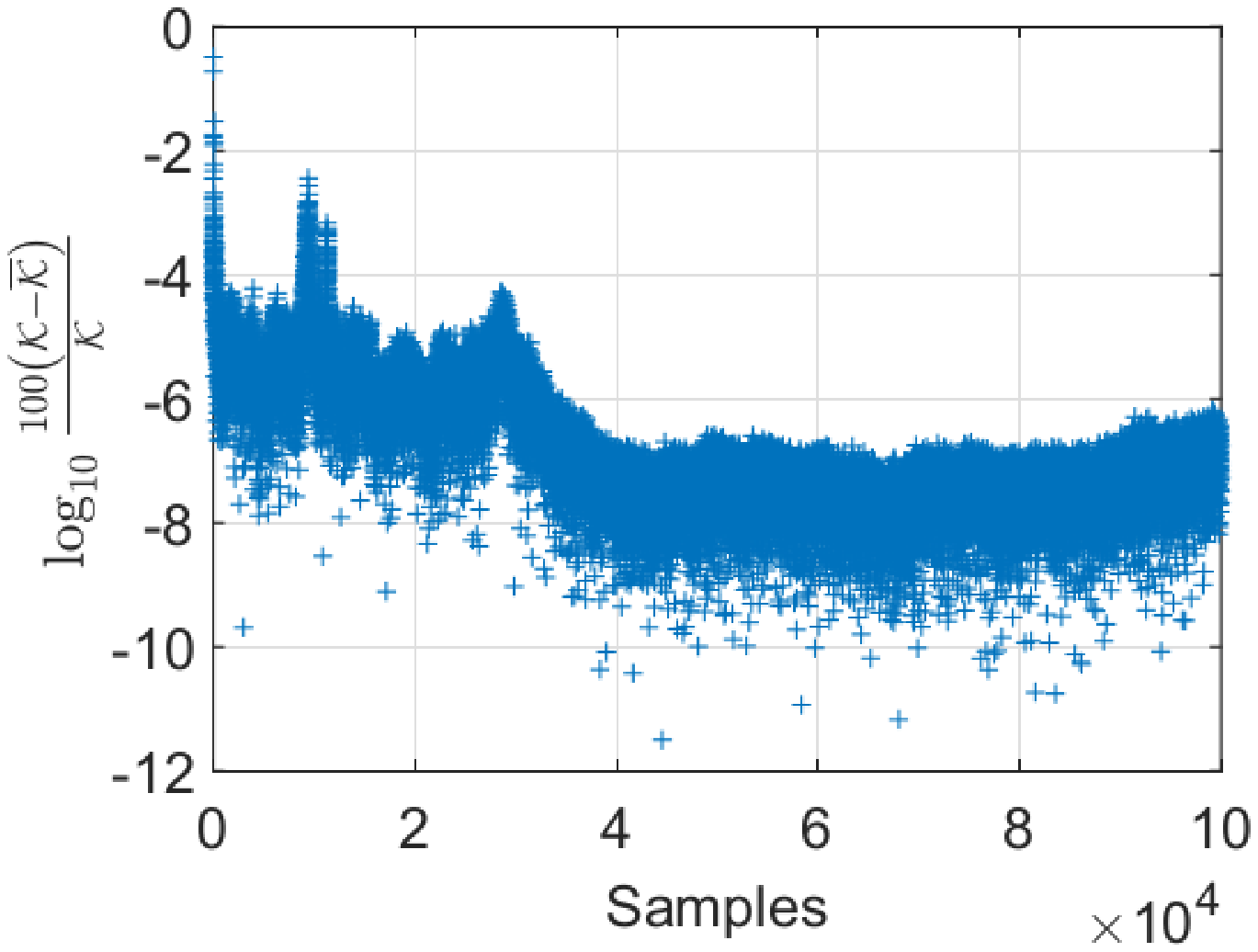}
\par\end{centering}
\label{fig:sampleErrors}}\subfloat[]{\begin{centering}
\includegraphics[bb=0bp 0bp 400bp 315bp,width=0.48\columnwidth]{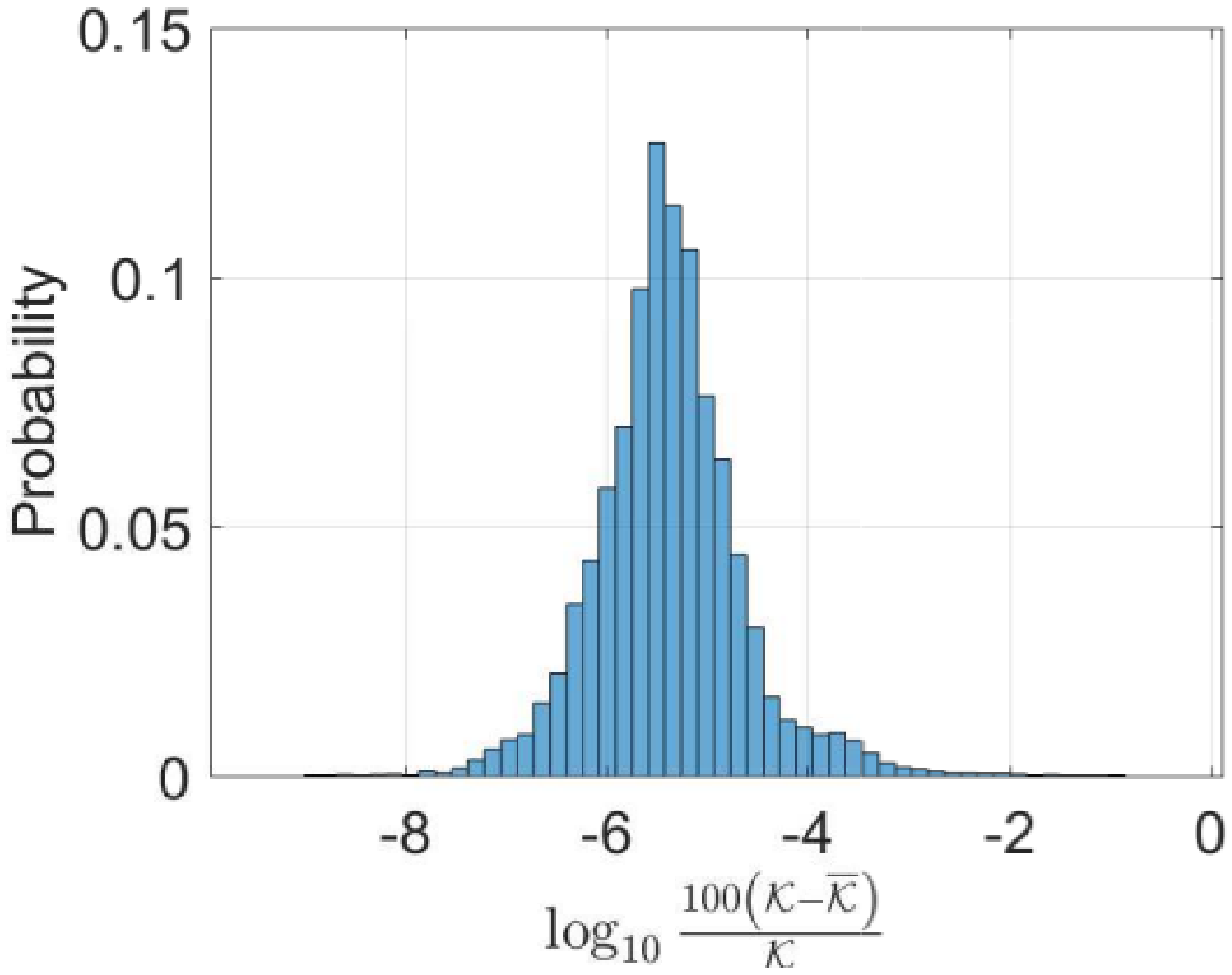}
\par\end{centering}
\label{fig:energyDistribution}}
\par\end{centering}
\caption{Energy difference between the integral and CoG-based modeling approaches
for a 10-section continuum arm. (a) Energy difference for $10^{6}$
randomly selected joint-space displacement and velocity samples, (b)
Histogram of the energy difference for the same samples.}
\label{fig:energyDiffAnalysis}
\end{figure}

In this section, we statistically validate the coefficients generated
in the previous section for a ten-section continuum robot model. For
an $n$ section continuum arm, $6n$ variables are required to compute
the kinetic energy ($3n$ joint-space displacements and velocities).
Assuming the $L_{0}=0.15\,\text{m}$, $r_{i}=0.0125\,\text{m}$, and
$m_{i}=0.1\,\text{kg}$ (physical parameters corresponding to the
prototype arm shown in Fig. \ref{fig:iitArm}), here we generate $10^{6}$
samples of uniformly distributed values within $\left[0,0.07\,\text{m}\right]$
and $\left[-L_{0},L_{0}\right]\,\text{ms}^{-1}$ for $\boldsymbol{q}$
and $\dot{\boldsymbol{q}}$ respectively for the continuum arm numerical
model. The difference of the complete system kinetic energies is computed
by taking the cumulative of section-wise energy differences given
by \eqref{eq:dKw_i} and \eqref{eq:dKv_i}. %
The energy difference percentages, normalized to $\text{max}\left(\mathcal{K}_{i}^{\upsilon}+\mathcal{K}_{i}^{\omega}\right)$,
for each sample, are then computed and plotted in Fig. \ref{fig:sampleErrors}.
Note that, $\text{max}\left(\mathcal{K}_{i}^{\upsilon}+\mathcal{K}_{i}^{\omega}\right)$,
is not the absolute maximum kinetic energy for the given robot, but
rather it is a statistical upper bound, and therefore the energy error
percentages computed here are conservative, and the actual error is
likely to be significantly lower in practice. The error percentage
distribution is shown in Fig. \ref{fig:energyDistribution}. %
The figure shows that the energy difference is essentially negligible
with $10^{-6}$ mean error percentage. %
The results show that the computed energy scalars are accurate and
applicable for arbitrary length continuum arms without undesirable
error propagation, eliminating the need for complex integral terms.

\subsection{Potential Energy of Continuum Sections\label{subsec:Potential-Energy-of}}

As reported in \cite{godage2016dynamics}, a continuum arm is subjected
to gravitational and elastic potential energies. Elastic potential
energy, given by $\mathcal{P}^{e}=\frac{1}{2}\boldsymbol{q}^{T}\mathbf{K}_{e}\boldsymbol{q}$,%
only depends on $\boldsymbol{q}$ and is therefore independent of
the modeling approach herein. The gravitational potential energy for
the integral and CoG-based model can be defined as $\mathcal{P}_{i}^{g}=\int m_{i}\boldsymbol{g}^{T}\boldsymbol{p}^{i}$
and $\overline{\mathcal{P}}_{i}^{g}=m_{i}\boldsymbol{g}^{T}\overline{\boldsymbol{p}}^{i}$
respectively. Note that, $\mathcal{P}_{i}^{g}$ %
does not contain products of integrable terms. Therefore, $\mathcal{P}_{i}^{g}$
can be simplified to $\mathcal{P}_{i}^{g}=m_{i}\boldsymbol{g}^{T}\left(\int\boldsymbol{p}^{i}\right)$
and from the definition \eqref{eq:HTM_c_ith}, then becomes $\mathcal{P}_{i}^{g}=m_{i}\boldsymbol{g}^{T}\left(\overline{\boldsymbol{p}}^{i}\right)=\overline{\mathcal{P}}_{i}^{g}$.
Thus, the gravitational potential energy is identical in both models.

\section{Recursive Formulation of Equations of Motion\label{sec:Recursive-Formulation-of}}

This section utilizes the energy relationships derived in Section
\ref{subsec:Finding-the-Scaling} to formulate the recursive form
of the EoM. %
Let the Lagrangian of the system using the CoG-based model be $\overline{\mathcal{K}}-\mathcal{\overline{P}}$.
Then the EoM in standard form is given by

\begin{align}
\overline{\mathbf{M}}\ddot{\boldsymbol{q}}+\overline{\mathbf{C}}\dot{\boldsymbol{q}}+\overline{\boldsymbol{G}} & =\boldsymbol{\tau}\label{eq:eom}
\end{align}
where $\overline{\mathbf{M}}\in\mathbb{R}^{3n\times3n}$, $\overline{\mathbf{C}}\in\mathbb{R}^{3n\times3n}$,
$\overline{\boldsymbol{G}}\in\mathbb{R}^{3n\times1}$, and $\tau\in\mathbb{R}^{3n\times1}$
are generalized inertia matrix, centrifugal and Coriolis force matrix,
conservative force matrix, and joint-space input force vector. 

From the theorems derived in \cite{godage2016dynamics}, we can%
{} decompose these matrices into section-wise contributions as $\overline{\mathbf{M}}=\sum\overline{\mathbf{M}}_{i}$,
$\overline{\mathbf{C}}=\sum\overline{\mathbf{C}}_{i}$, and $\overline{\boldsymbol{G}}=\sum\overline{\boldsymbol{G}}_{i}$
respectively. In this section, we derive the section-wise contributions
in recursive form to compute the EoM in \eqref{eq:eom}.

\subsection{Generalized Inertia Matrix $\left(\overline{\mathbf{M}}_{i}\right)$\label{subsec:Generalized-Inertia-Matrix}}

Analogous to the integral modeling approach \cite{godage2016dynamics},%
{} we can define the $i^{th}$ section kinetic energy to be the sum
of the scaled (using the energy scalars to math the integral model)
angular and linear kinetic energies, $\overline{\mathcal{K}}_{i}=\overline{\mathcal{K}}_{i}^{\upsilon}+\overline{\mathcal{K}}_{i}^{\omega}$.
Thus, by applying the partial derivatives with respect to the joint-space
velocities on $\overline{\mathcal{K}}_{i}$, we obtain the generalized
inertia matrix contributions as%
, $\overline{\mathbf{M}}_{i}=\mathbf{\overline{M}}_{i}^{\omega}+\overline{\mathbf{M}}_{i}^{\upsilon}$.
Using the angular velocity Jacobian, $\overline{\mathbf{J}}_{i}^{\Omega}$
and %
the scalar coefficients derived in Section \ref{subsec:Finding-the-Scaling},
we can derive $\mathbf{\overline{M}}_{i}^{\omega}$ as%

\begin{align}
\mathbf{\overline{M}}_{i}^{\omega} & =I_{xx}\mathbb{T}_{2}\left[\begin{array}{cc}
\beta_{1}^{\omega}\sigma_{11}^{\omega} & \beta_{2}^{\omega}\sigma_{12}^{\omega}\\
\beta_{2}^{\omega}{\sigma_{12}^{\omega}}^{T} & \beta_{3}^{\omega}\sigma_{22}^{\omega}
\end{array}\right]\label{eq:Mw_c_scaled}
\end{align}
where $\sigma_{11}^{\omega}=\left(\mathbf{J}_{i-1}^{\Omega}\overline{\mathbf{R}}_{i}\right)^{T}\mathbf{J}_{i-1}^{\Omega}\overline{\mathbf{R}}_{i}$,
$\sigma_{12}^{\omega}=\left(\mathbf{J}_{i-1}^{\Omega}\overline{\mathbf{R}}_{i}\right)^{T}\overline{\mathbf{R}}_{i,\boldsymbol{q}_{i}^{T}}$,
and $\sigma_{22}^{\omega}=\overline{\mathbf{R}}_{i,\boldsymbol{q}_{i}^{T}}^{T}\overline{\mathbf{R}}_{i,\boldsymbol{q}_{i}^{T}}$
respectively. 

Equivalently, by applying the recursive form of the Jacobian in \eqref{eq:Jv_c}
and the energy scalars derived in Section \ref{subsec:Finding-the-Scaling},
we can derive $\mathbf{\overline{M}}_{i}^{\upsilon}$ as

\begin{align}
\mathbf{\overline{M}}_{i}^{\upsilon} & =m_{i}\left[\begin{array}{cc}
\sigma_{11}^{\upsilon} & \sigma_{12}^{\upsilon}\\
{\sigma_{12}^{\upsilon}}^{T} & \sigma_{22}^{\upsilon}
\end{array}\right]\label{eq:Mv_c}
\end{align}
where $\sigma_{11}^{\upsilon}={\mathbf{J}_{i-1}^{\upsilon}}^{T}\left(\mathbf{J}_{i-1}^{\upsilon}+2\mathbf{J}_{i-1}^{\Omega}\overline{\boldsymbol{p}}_{i}\right)$$+\beta_{1}^{\upsilon}\left(\mathbf{J}_{i-1}^{\Omega}\overline{\boldsymbol{p}}_{i}\right)^{T}\mathbf{J}_{i-1}^{\Omega}\overline{\boldsymbol{p}}_{i}$,
$\sigma_{12}^{\upsilon}=\left(\mathbf{J}_{i-1}^{\upsilon}+\beta_{2}^{\upsilon}\mathbf{J}_{i-1}^{\Omega}\overline{\boldsymbol{p}}_{i}\right)^{T}\overline{\boldsymbol{p}}_{i,\boldsymbol{q}_{i}^{T}}$,
and $\sigma_{22}^{\upsilon}=\beta_{3}^{\upsilon}\overline{\boldsymbol{p}}_{i,\boldsymbol{q}_{i}^{T}}^{T}\overline{\boldsymbol{p}}_{i,\boldsymbol{q}_{i}^{T}}$.%

\subsection{Coriolis and Centrifugal Force Matrix $\left(\overline{\mathbf{C}}_{i}\right)$}

Using partial derivatives of $\overline{\mathbf{M}}_{i}$, the Christoffel
symbols of the $2^{nd}$ kind are used to derive the $\mathbf{\overline{C}}_{i}$
elements as

\begin{align}
\left[\mathbf{\overline{C}}_{i}\right]_{jk} & =\frac{1}{2}\sum_{h=1}^{3i}\!\!\left(\left[\overline{\mathbf{M}}_{i}\right]_{kj,q_{h}}\!\!+\!\left[\mathbf{\overline{M}}_{i}\right]_{kh,q_{j}}\!\!-\!\left[\overline{\mathbf{M}}_{i}\right]_{hj,q_{k}}\right)\dot{q}_{h}\label{eq:Ckj}
\end{align}

Noting that $\overline{\mathbf{M}}_{i}=\mathbf{\overline{M}}_{i}^{\omega}+\overline{\mathbf{M}}_{i}^{\upsilon}$,
by applying partial derivatives with respect to $h\in\boldsymbol{q}^{i}$,
we get $\overline{\mathbf{M}}_{i,h}=\mathbf{\overline{M}}_{i,h}^{\omega}+\overline{\mathbf{M}}_{i,h}^{\upsilon}$.
Hence, considering the variable with respect to which the partial
derivation is carried out, we can obtain $\mathbf{\overline{M}}_{i,h}^{\omega}$
as %

\begin{align}
\mathbf{\overline{M}}_{i,h}^{\omega} & =I_{xx}\mathbb{T}_{2}\begin{cases}
\left[\begin{array}{cc}
\eta_{11}^{\omega} & \eta_{12}^{\omega}\\
{\eta_{12}^{\omega}}^{T} & \eta_{22}^{\omega}
\end{array}\right] & ;\:h\in\boldsymbol{q}^{i-1}\\
\left[\begin{array}{cc}
\gamma_{11}^{\omega} & \gamma_{12}^{\omega}\\
{\gamma_{12}^{\omega}}^{T} & \gamma_{22}^{\omega}
\end{array}\right] & ;\:h\in\boldsymbol{q}_{i}
\end{cases}\label{eq:dMw_C}
\end{align}
where $\left(\mathbf{H}_{i-1}^{\Omega}\right)_{h}=\mathbf{J}_{i-1,h}^{\Omega}$
is the submatrix of $\mathbf{H}_{i-1}^{\Omega}$ and the terms are
listed in Tab. \ref{tab:dMW_C}.%

\begin{table}[tb]
\caption{Terms associated with \eqref{eq:dMw_C} and \eqref{eq:dMv_c}.}
\label{tab:dMW_C}
\centering{}%
\begin{tabular}{>{\raggedright}p{0.95\columnwidth}}
$\eta_{11}^{\omega}=2\beta_{1}^{\omega}\left(\mathbf{J}_{i-1}^{\Omega}\overline{\mathbf{R}}_{i}\right)^{T}\left(\mathbf{H}_{i-1}^{\Omega}\right)_{h}\overline{\mathbf{R}}_{i}$\tabularnewline
$\eta_{12}^{\omega}=\beta_{2}^{\omega}\left\{ \left(\mathbf{H}_{i-1}^{\Omega}\right)_{h}\overline{\mathbf{R}}_{i,\boldsymbol{q}_{i}}\right\} ^{T}\overline{\mathbf{R}}_{i}$\tabularnewline
$\eta_{11}^{\omega}=\boldsymbol{0}$\tabularnewline
$\gamma_{11}^{\omega}=2\beta_{1}^{\omega}\left(\mathbf{J}_{i-1}^{\Omega}\overline{\mathbf{R}}_{i,h}\right)^{T}\mathbf{J}_{i-1}^{\Omega}\overline{\mathbf{R}}_{i}$\tabularnewline
\multirow{2}{0.95\columnwidth}{$\gamma_{12}^{\omega}=\beta_{2}^{\omega}\left(\mathbf{J}_{i-1}^{\Omega}\overline{\mathbf{R}}_{i,\boldsymbol{q}_{i},h}\right)^{T}\overline{\mathbf{R}}_{i}$$+\left(\mathbf{J}_{i-1}^{\Omega}\overline{\mathbf{R}}_{i,\boldsymbol{q}_{i}}\right)^{T}\overline{\mathbf{R}}_{i,h}$}\tabularnewline
\tabularnewline
$\gamma_{22}^{\omega}=2\beta_{3}^{\omega}\overline{\mathbf{R}}_{i,\boldsymbol{q}_{i},h}^{T}\overline{\mathbf{R}}_{i,\boldsymbol{q}_{i}^{T}}$\tabularnewline
$\eta_{11}^{\upsilon}=2\left({\mathbf{H}_{i-1}^{\upsilon}}\right)_{h}^{T}\left(\mathbf{J}_{i-1}^{\upsilon}+\mathbf{J}_{i-1}^{\Omega}\overline{\boldsymbol{p}}_{i}\right)+2{\mathbf{J}_{i-1}^{\upsilon}}^{T}\left(\mathbf{H}_{i-1}^{\Omega}\right)_{h}\overline{\boldsymbol{p}}_{i}+2\beta_{1}^{\upsilon}\left(\mathbf{J}_{i-1}^{\Omega}\overline{\boldsymbol{p}}_{i}\right)^{T}\left(\mathbf{H}_{i-1}^{\Omega}\right)_{h}\overline{\boldsymbol{p}}_{i}$\tabularnewline
${\eta_{12}^{\upsilon}}=\left\{ \left(\mathbf{H}_{i-1}^{\upsilon}\right)_{h}+\beta_{2}^{\upsilon}\mathbf{H}_{i-1}^{\Omega}\overline{\boldsymbol{p}}_{i}\right\} ^{T}\overline{\boldsymbol{p}}_{i,\boldsymbol{q}_{i}^{T}}$\tabularnewline
$\eta_{22}^{\upsilon}=\boldsymbol{0}$\tabularnewline
$\gamma_{11}^{\upsilon}=2{\mathbf{J}_{i-1}^{\upsilon}}^{T}\mathbf{J}_{i-1}^{\Omega}\overline{\boldsymbol{p}}_{i,h}+2\beta_{1}^{\upsilon}\left(\mathbf{J}_{i-1}^{\Omega}\overline{\boldsymbol{p}}_{i,h}\right)^{T}\mathbf{J}_{i-1}^{\Omega}\overline{\boldsymbol{p}}_{i}$\tabularnewline
$\gamma_{12}^{\upsilon}=\left(\mathbf{J}_{i-1}^{\upsilon}+\beta_{2}^{\upsilon}\mathbf{J}_{i-1}^{\Omega}\overline{\boldsymbol{p}}_{i}\right)^{T}\overline{\boldsymbol{p}}_{i,\boldsymbol{q}_{i}^{T},h}+\beta_{2}^{\upsilon}\left(\mathbf{J}_{i-1}^{\Omega}\overline{\boldsymbol{p}}_{i,h}\right)^{T}\overline{\boldsymbol{p}}_{i,\boldsymbol{q}_{i}^{T}}$\tabularnewline
$\gamma_{22}^{\upsilon}=2\beta_{3}^{\upsilon}\overline{\boldsymbol{p}}_{i,\boldsymbol{q}_{i}^{T},h}^{T}\overline{\boldsymbol{p}}_{i,\boldsymbol{q}_{i}^{T}}$\vspace{1mm}\tabularnewline
\hline 
\end{tabular}
\end{table}

Similarly, the $\mathbf{\overline{M}}_{i,h}^{\upsilon}$ is given
by 

\begin{align}
\mathbf{\overline{M}}_{i,h}^{\upsilon} & =m_{i}\begin{cases}
\left[\begin{array}{cc}
\eta_{11}^{\upsilon} & \eta_{12}^{\upsilon}\\
{\eta_{12}^{\upsilon}}^{T} & \eta_{22}^{\upsilon}
\end{array}\right] & ;\:h\in\boldsymbol{q}^{i-1}\\
\left[\begin{array}{cc}
\gamma_{11}^{\upsilon} & \gamma_{12}^{\upsilon}\\
{\gamma_{12}^{\upsilon}}^{T} & \gamma_{22}^{\upsilon}
\end{array}\right] & ;\:h\in\boldsymbol{q}_{i}
\end{cases}\label{eq:dMv_c}
\end{align}
where $\left(\mathbf{H}_{i-1}^{\upsilon}\right)_{h}=\left(\mathbf{J}_{i-1}^{\upsilon}\right)_{,h}$
is the submatrix of $\mathbf{H}_{i-1}^{\upsilon}$ and the terms are
listed in Tab. \ref{tab:dMW_C}.%

\subsection{Conservative Force Matrix $\left(\mathbf{G}_{i}\right)$}

The gravitational $\left(\mathcal{P}_{i}^{g}\right)$ and elastic
$\left(\mathcal{P}_{i}^{e}\right)$ potential energy contributes to
the total potential energy of a continuum section, $\mathcal{P}_{i}=\mathcal{P}_{i}^{g}+\mathcal{P}_{i}^{e}$.
Therefore, the contribution to $\boldsymbol{G}_{i}$ can be written
as $\boldsymbol{G}_{i}=\boldsymbol{G}_{i}^{g}+\boldsymbol{G}_{i}^{e}$
\cite{godage2016dynamics}.%
{} The gravitational potential energy of the $i^{th}$ section can be
written as $\mathcal{P}_{i}^{g}=m_{i}\boldsymbol{g}^{T}\overline{\boldsymbol{p}}^{i}$.
As there are no products of integrable terms, $\boldsymbol{G}_{i}^{g}$
is identical for both integral and CoG-based dynamic models and can
be directly derived as 

\begin{align}
{\boldsymbol{G}_{i}^{g}}^{T} & =\left(m_{i}\boldsymbol{g}^{T}\overline{\boldsymbol{p}}^{i}\right)_{,\left(\boldsymbol{q}^{i}\right)^{T}}\nonumber \\
 & =m_{i}\boldsymbol{g}^{T}{\mathbf{R}^{i-1}}\left[\begin{array}{cc}
\mathbf{J}_{i-1}^{\upsilon}+\mathbf{J}_{i-1}^{\Omega}\overline{\boldsymbol{p}}_{i} & \overline{\boldsymbol{p}}_{i,\boldsymbol{q}_{i}^{T}}\end{array}\right]_{,}\label{eq:Gi_c}
\end{align}
where the derivation is included in Appendix \ref{a:Conservative-Force-Vector,a}.

The elastic potential energy, $\mathcal{P}^{e}=\frac{1}{2}\boldsymbol{q}^{T}\mathbf{K}_{e}\boldsymbol{q}$,
is independent of mass or the relative position in the task-space.
Hence, similar to $\boldsymbol{G}_{i}^{g}$, $\boldsymbol{G}_{i}^{e}$
identical in both integral and CoG-based systems and could be readily
formulated as%

\begin{align}
\boldsymbol{G}_{i}^{e} & =\mathcal{P}_{i,\boldsymbol{q}_{i}}^{e}=\mathbf{K}_{e}\boldsymbol{q}_{i}\label{eq:Ge}
\end{align}

\subsection{Numerical Simulation Model}

The EoM numerical model was implemented in Matlab 2017a. The HTM was
implemented in Maple 16 \cite{Maplesoft:2010:MUM} symbolically and
manipulated to derive the CoG-based terms and the partial derivatives
thereof. Similarly, the kinematic terms used for computing the forward
kinematics and related terms (Jacobians given by \eqref{eq:Jo}, \eqref{eq:Jv}
and Hessians given by \eqref{eq:Ho}, \eqref{eq:Hv}) were computed
by making $\xi_{i}=1$ of terms related to the $i^{th}$ section.
These results were then implemented as Matlab functions. 

\begin{figure}[tb]
\begin{centering}
\subfloat[]{\begin{raggedleft}
\includegraphics[width=0.47\columnwidth]{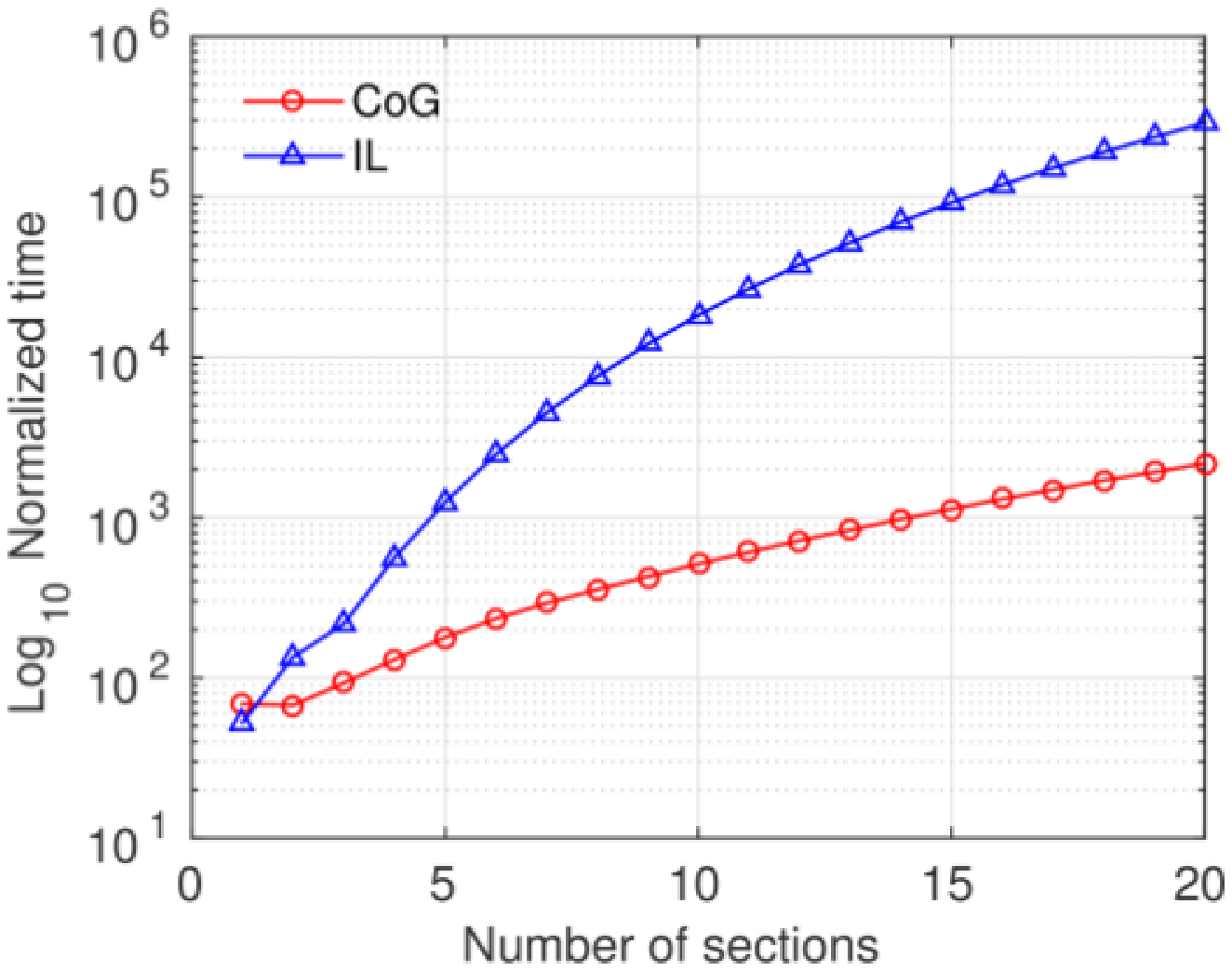}
\par\end{raggedleft}
\label{fig:AlgComplexity}}\subfloat[]{\begin{raggedright}
\includegraphics[width=0.47\columnwidth]{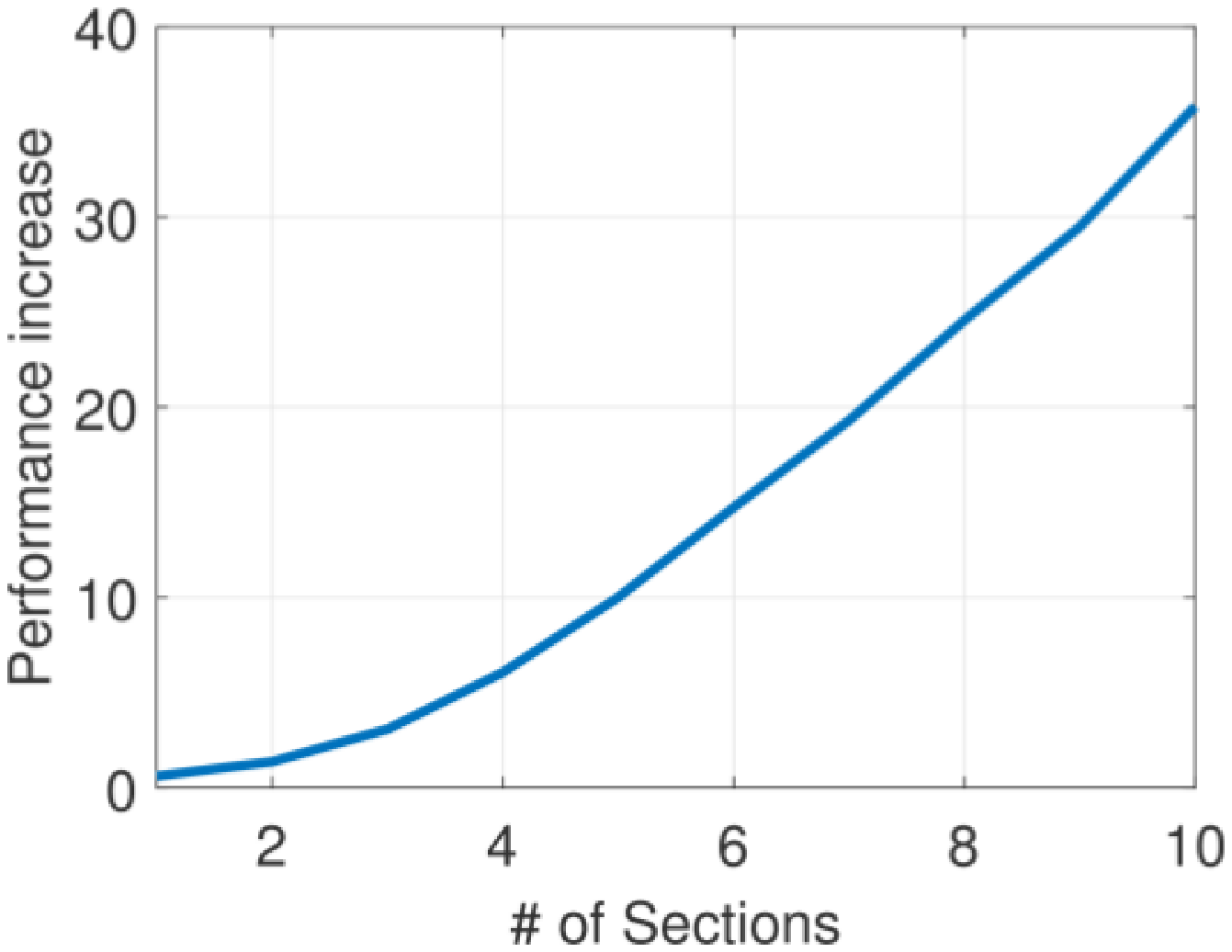}
\par\end{raggedright}
\label{fig:perfIncrease}}
\par\end{centering}
\caption{(a) Time complexity comparison and (b) performance gain between integral
and CoG-based dynamics as the number of continuum sections (i.e.,
DoF) increase. }
\end{figure}

The recursive algorithm shown in Algorithm \ref{alg:eom} was implemented
in Matlab Simulink environment is used to numerically solve the EoM
using the integrated ODE15s solver. Figure \ref{fig:AlgComplexity}
compares the CoG-based model against the integral dynamics model reported
in \cite{godage2016dynamics} where the former is of $\mathcal{O}\left(n^{2}\right)$
whereas the latter is $\mathcal{O}\left(n^{3}\right)$. For a single
section system (three DoF), both models show similar computation cost,
but the numerical efficiency of the proposed model is evident for
multisection continuum arms. The performance gain achieved by the
proposed model relative to the integral dynamics model is plotted
in Fig. \ref{fig:perfIncrease}. It can be seen that the CoG-model
is ideally suited for simulating dynamics of multisection continuum
arms. The dynamic parameters and coefficients, such as $\mathbf{K}_{i}^{e}$
and $\mathbf{D}_{i}$, are difficult to measure or accurately estimate
solely through physical and material properties. Therefore such parameters
were identified through an iterative system characterization process.
The reader is referred to \cite{godage2016dynamics} for a detailed
discussion of the process including the information regarding the
experimental setup and continuum arm shape measurement techniques.%

\begin{algorithm}[tb]
\noindent\fbox{\begin{minipage}[t]{1\columnwidth - 2\fboxsep - 2\fboxrule}%
\textbf{FOR} i \textbf{FROM} 1 \textbf{TO} n \textbf{DO}

$\qquad$compute $\boldsymbol{p}_{i}$, $\mathbf{R}_{i}$, $\boldsymbol{\overline{p}}_{i}$,
$\overline{\mathbf{R}}_{i}$, and partial derivatives

$\qquad$compute $\mathbf{M}_{i}=\mathbf{M}_{i}+(\mathbf{M}_{i}^{\omega}+\mathbf{M}_{i}^{\upsilon})$

$\qquad$compute $\boldsymbol{G}_{i}=\boldsymbol{G}_{i}+(\boldsymbol{G}_{i}^{p}+\boldsymbol{G}_{i}^{e})$

$\qquad$\textbf{FOR} h \textbf{FROM} 1 \textbf{TO} n \textbf{DO}

$\qquad$$\qquad$compute $\mathbf{M}_{i,h}$

$\qquad$update $\mathbf{J}_{i}^{\upsilon}$,$\mathbf{J}_{i}^{\omega}$,
$\boldsymbol{p^{i}}$, and $\mathbf{R}^{i}$

\textbf{FOR} i \textbf{FROM} 1 \textbf{TO} n \textbf{DO}

$\qquad$compute $\mathbf{C}_{i}$= $f\left(\mathbf{M}_{i,h}\right)$

\textbf{SOLVE}

$\qquad$$\overline{\mathbf{M}}\ddot{\boldsymbol{q}}+\left(\mathbf{\overline{C}}+\mathbf{D}\right)\dot{\boldsymbol{q}}+\overline{\boldsymbol{G}}=\boldsymbol{\tau}$ %
\end{minipage}}

\caption{Outline of the CoG-based EoM derivation via recursive Lagrangian formulation.}
\label{alg:eom}
\end{algorithm}

\section{Comparison to Experimental Results and Integral Dynamics\label{sec:Experimental-Results}}

The prototype continuum arm utilized in the following experiments
is shown in Fig. \ref{fig:iitArm}. %
This section uses the same experimental data reported in \cite{godage2016dynamics}
and compares the proposed CoG-based dynamics against the integral
dynamics and the experimental results therein. 

\begin{figure}[tb]
\begin{centering}
\includegraphics[width=1\columnwidth]{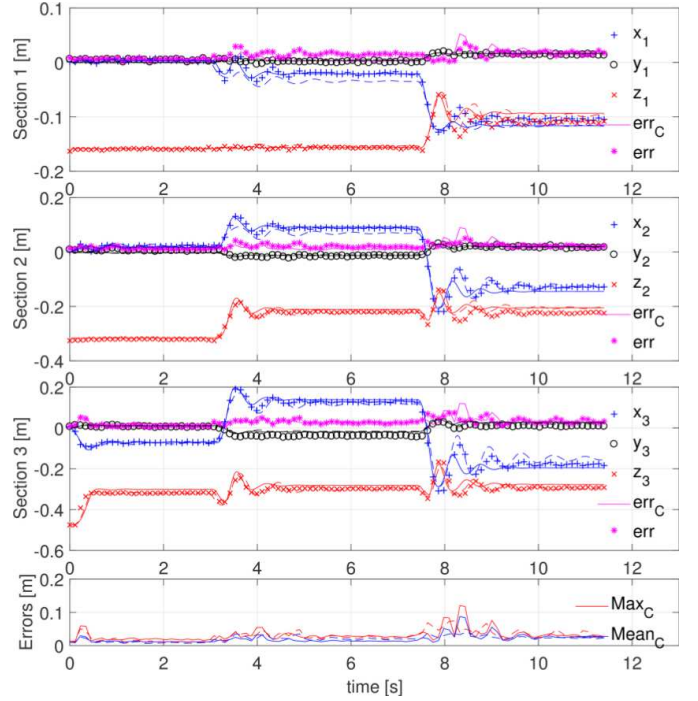}
\par\end{centering}
\caption{Tip coordinate (X: blue,Y: black, and Z: red) trajectories of continuum
sections for the first experiment. Experimental data are denoted by
$\boldsymbol{\circ,+,\times}$ marks respectively for each section.
Integral dynamic results are shown in colored dashed (-~-) lines
where as CoG-based dynamics are shown in solid lines of the same colors.
Position errors (the Euclidean distance) at each tip are shown in
magenta color for integral dynamics (dashed, -~-) and CoG dynamics
(solid) lines are also included. }
\label{fig:exp_1}
\end{figure}

The first experiment involves section-wise actuation of all the sections
on the $y=0$ plane. The joint-space variables (physically the PMA's
of continuum sections), $l_{33}$, $l_{22}$, and $l_{11}$, are supplied
with $600\,kPa$, $500\,kPa$, and $500\,kPa$ step pressure inputs
at $t=0\,s$, $t=3.2s$, and $t=7.55s$ respectively. The EoM given
in \eqref{eq:eom}, derived using the proposed CoG-based approach,
is the provided the same pressure input to simulate the forward dynamics.
The simulation took $1.13\,s$ to complete, which is 6.69 times faster
than real-time. The resultant joint-space trajectories are then applied
to the kinematic model given by \eqref{eq:eom} to compute the associated
tip task-space trajectories. The section tip coordinate task-space
trajectories, measured using a two-camera setup \cite{godage2016dynamics}
(illustrated in various shaped discrete markers) are then compared
to the simulated task-space trajectories (drawn in solid lines) in
Fig. \ref{fig:exp_1}. In addition, the task-space trajectories computed
by the integral dynamics \cite{godage2016dynamics} are also included
to compare the performance of the two approaches (shown in dotted
lines). The errors between the experimental data versus CoG-based
model and integral dynamics are also shown in each of the subplots
for ease of comparison. It can be seen that the difference in errors
and simulated results between the two numerical models are negligible.
The aggregated error, plotted in the bottom subplot shows the maximum
error among the three tip positions and the mean of the position errors
of all sections. It can be seen that the proposed model matches the
integral dynamics proposed in \cite{godage2016dynamics}. Similar
plots are generated for two further experiments detailed below.

\begin{figure}[tb]
\begin{centering}
\includegraphics[width=1\columnwidth]{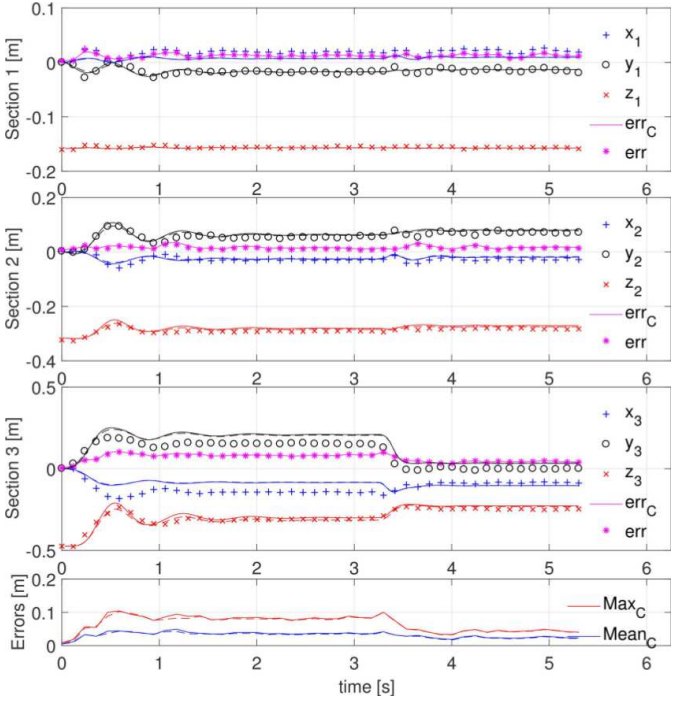}
\par\end{centering}
\caption{Tip coordinate (X: blue,Y: black, and Z: red) trajectories of continuum
sections for the second experiment. The legend is the same as Fig.
\ref{fig:exp_1}.}
\label{fig:exp_2}
\end{figure}

The second experiment involves the actuation of the distal and mid
section in two, non-parallel bending planes while the base section
remains unactuated. Step pressure inputs of $300\,kPa$ and $500\,kPa$
were applied to $l_{23}$ at $t=0\,s$ and $l_{33}$ at $t=3.3\,s$.
The resulting experimental and simulated task-space trajectories (using
both integral dynamics and CoG-based dynamics) are shown in Fig. \ref{fig:exp_2}.
The base section, though unactuated deforms passively to balance the
dynamic forces induced by the other moving sections, which is correctly
modeled by both integral and CoG-based dynamic models. The numerical
computation was 7.8 times faster than real-time and completed within
$0.89\,s$. Both models show comparable errors during the transient
phase of the step response, but both models correctly simulate the
steady-state dynamics afterwards. The error in this experiment also
varies during the step input transient stages, but section settles
down quickly.

\begin{figure}[tb]
\begin{centering}
\includegraphics[width=1\columnwidth]{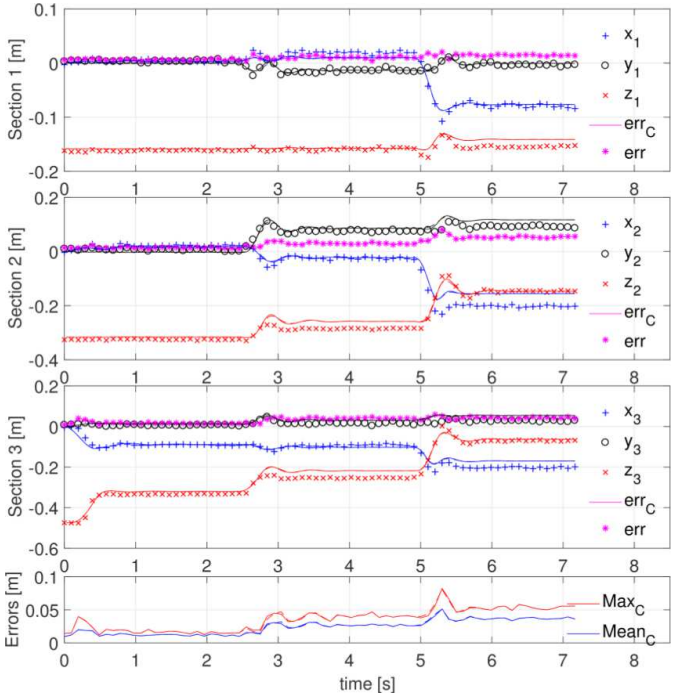}
\par\end{centering}
\caption{Tip coordinate (X: blue,Y: black, and Z: red) trajectories of continuum
sections for the third experiment. The legend is the same as Fig.
\ref{fig:exp_1}.}
\label{fig:exp_3}
\end{figure}

The third experiment extends the second and includes the actuation
of the base section. The prototype and the dynamic model are provided
pressure step inputs of $500\,kPa$, $300\,kPa$, and $300\,kPa$
are respectively to actuators $l_{33}$ at $t=0\,s$ , $l_{23}$ at
$t=2.55\,s$ and $l_{11}$ at $t=5.05~s$ and maintained during the
experiment, and cause the continuum arm sections to deform in non-parallel
planes. Figure \ref{fig:exp_3} compares the integral and CoG-based
dynamics against the experimental results reported in \cite{godage2016dynamics}.
The simulation only took $1.3\,s$ to complete this $7.9\,s$ long
experiment, which is 7.3 times faster than real-time. It can be seen
that the CoG-based dynamics agrees with both the integral dynamics
and experimental results. These experimental and empirical data demonstrate
that the proposed, numerically efficient CoG-based dynamic model for
variable-length multisection continuum arms successfully simulates
both the transient and steady-state dynamic behaviors well.

\section{Conclusions\label{sec:Conclusions}}

Multisection continuum arms have strong potential for use in human-friendly
spaces. Despite continued research, they have yet to make their mark
outside the laboratory settings. A key reason for this is the lack
of numerically efficient dynamic models that can be used in sub real-time.
Accuracy, numerical stability, and efficiency are critical for dynamic
models to be used in dynamic control. Limited research has been conducted
on physically accurate dynamic modeling of multisection continuum
arms experimental validation thereof. In this paper, a novel, CoG-based
dynamic model was proposed. The work extended our prior work on CoG-based
modeling of a single continuum section to derive a general model that
can be used not only on arbitrarily long continuum arms but also such
robots of varying physical sizes. The results show that the model
accommodates arbitrarily long variable-length multisection continuum
arms and various length-radii combinations, considers both linear
and angular kinetic energies at the CoGs of sections for more accuracy
in energy computation, matches energy through a series of constant
(for any variable-length multisection continuum arm) energy shaping
coefficients, derives the EoM terms recursively, attains $\mathcal{O}\left(n^{2}\right)$
complexity for continuous (non-discretized) dynamic model for variable-length
arms, and is 6-8 times numerically efficient than real-time for a
three-section continuum arm model (suitable for implementing dynamic
control schemes) and runs at 9.5~kHz. The model was experimentally
validated on a three-section continuum arm and showed that results
agree well with both the robot output as well as the integral dynamic
models.

\appendices{}

\section{Mathematical Derivations\label{asec:Mathematical-Derivations}}

\subsection{Recursive Kinematic Relationships for any $i^{th}$ Continuum Section\label{asubsec:Recursive-Kinematic-Relationship}}

\subsubsection{Angular Body Velocity\label{a:Oi_derivation}}

\begin{align}
\boldsymbol{\Omega}_{i} & ={\mathbf{R}^{i}}^{T}\dot{\mathbf{R}}^{i}\nonumber \\
 & =\left(\mathbf{R}^{i-1}\mathbf{R}_{i}\right)^{T}\left(\dot{\mathbf{R}}^{i-1}\mathbf{R}_{i}+\mathbf{R}^{i-1}\mathbf{\dot{R}}_{i}\right)\nonumber \\
 & =\mathbf{R}_{i}\left\{ \left({\mathbf{R}^{i-1}}^{T}\dot{\mathbf{R}}^{i-1}\right)\mathbf{R}_{i}+\left({\mathbf{R}^{i-1}}^{T}\mathbf{R}^{i-1}\right)\mathbf{\dot{R}}_{i}\right\} \nonumber \\
 & =\mathbf{R}_{i}^{T}\left(\boldsymbol{\Omega}_{i-1}\mathbf{R}_{i}+\dot{\mathbf{R}}_{i}\right)\label{eq:Oi_derivation}
\end{align}

\subsubsection{Linear body Velocity\label{a:vi_derivation}}

\begin{align}
\boldsymbol{\upsilon}_{i} & ={\mathbf{R}^{i}}^{T}\dot{\boldsymbol{p}}^{i}\nonumber \\
 & =\left(\mathbf{R}^{i-1}\mathbf{R}_{i}\right)^{T}\left(\dot{\boldsymbol{p}}^{i-1}+\dot{\mathbf{R}}^{i-1}\mathbf{R}_{i}+\mathbf{R}^{i-1}\boldsymbol{\dot{p}}_{i}\right)\nonumber \\
 & =\mathbf{R}_{i}^{T}\left\{ \left({\mathbf{R}^{i-1}}^{T}\dot{\boldsymbol{p}}^{i-1}\right)\mathbf{R}_{i}+\left({\mathbf{R}^{i-1}}^{T}\dot{\mathbf{R}}^{i-1}\right)\boldsymbol{p}_{i}\cdots\right.\nonumber \\
 & \qquad\left.+\left({\mathbf{R}^{i-1}}^{T}\mathbf{R}^{i-1}\right)\dot{\boldsymbol{p}}_{i}\right\} \nonumber \\
 & =\mathbf{R}_{i}^{T}\left(\boldsymbol{v}_{i-1}+\boldsymbol{\Omega}_{i-1}\boldsymbol{p}_{i}+\dot{\boldsymbol{p}}_{i}\right)\label{eq:Vi_derivation}
\end{align}

\subsubsection{Angular Body Velocity Jacobian\label{a:Joi_derivation}}

\begin{align}
\mathbf{J}_{i}^{\Omega} & =\boldsymbol{\Omega}_{i,\left({\dot{\boldsymbol{q}}^{i}}\right)^{T}}\nonumber \\
 & =\mathbf{R}_{i}^{T}\left(\boldsymbol{\Omega}_{i-1}\mathbf{R}_{i}+\dot{\mathbf{R}}_{i}\right)_{,\left({\dot{\boldsymbol{q}}^{i}}\right)^{T}}\nonumber \\
 & =\mathbf{R}_{i}^{T}\left[\begin{array}{c|c}
\boldsymbol{\Omega}_{i-1,\left({\dot{\boldsymbol{q}}^{i-1}}\right)^{T}}\mathbf{R}_{i} & \dot{\mathbf{R}}_{i,\dot{\boldsymbol{q}}_{i}^{T}}\end{array}\right]\nonumber \\
 & =\mathbf{R}_{i}^{T}\left[\begin{array}{c|c}
\mathbf{J}_{i-1}^{\Omega}\mathbf{R}_{i} & \mathbf{R}_{i,\boldsymbol{q}_{i}}\end{array}\right]\label{eq:Jo_derivation}
\end{align}

\subsubsection{Linear Body Velocity Jacobian\label{a:Jvi_derivation}}

\begin{align}
\mathbf{J}_{i}^{\upsilon} & =\boldsymbol{\upsilon}_{i,\left(\dot{\boldsymbol{q}}^{i}\right)^{T}}\nonumber \\
 & =\mathbf{R}_{i}^{T}\left(\boldsymbol{\upsilon}_{i-1}+\boldsymbol{\Omega}_{i-1}\boldsymbol{p}_{i}+\dot{\boldsymbol{p}}_{i}\right){}_{,\left(\dot{\boldsymbol{q}}^{i}\right)^{T}}\quad\text{from \eqref{eq:wvi_recursive}}\nonumber \\
 & =\mathbf{R}_{i}^{T}\left[\begin{array}{c|c}
\boldsymbol{\upsilon}_{i-1,\left(\dot{\boldsymbol{q}}^{i-1}\right)^{T}}+\boldsymbol{\Omega}_{i-1,\left(\dot{\boldsymbol{q}}^{i-1}\right)^{T}}\boldsymbol{p}_{i} & \dot{\boldsymbol{p}}_{i,\dot{\boldsymbol{q}}_{i}^{T}}\end{array}\right]\nonumber \\
 & =\mathbf{R}_{i}^{T}\left[\begin{array}{c|c}
\mathbf{J}_{i-1}^{\upsilon}+\mathbf{J}_{i-1}^{\Omega}\boldsymbol{p}_{i} & \boldsymbol{p}_{i,\boldsymbol{q}_{i}^{T}}\end{array}\right]\label{eq:Jv_derivation}
\end{align}

\subsubsection{Angular Body Velocity Hessian\label{a:Hoi_derivation}}

\begin{align}
\mathbf{H}_{i}^{\Omega} & =\mathbf{J}_{i,\boldsymbol{q}^{i}}^{\Omega}\nonumber \\
 & =\left(\mathbf{R}_{i}^{T}\left[\begin{array}{cc}
\mathbf{J}_{i-1}^{\Omega}\mathbf{R}_{i} & \mathbf{R}_{i,\boldsymbol{q}_{i}^{T}}\end{array}\right]\right){}_{,\boldsymbol{q}^{i}}\nonumber \\
 & =\left[\!\!\begin{array}{c|c}
\mathbf{R}_{i}^{T}\left(\mathbf{J}_{i-1,\boldsymbol{q}^{i-1}}^{\Omega}\right)\mathbf{R}_{i} & \mathbf{R}_{i,\boldsymbol{q}_{i}^{T},\boldsymbol{q}^{i-1}}\\
\hline \mathbf{R}_{i,\boldsymbol{q}_{i}}^{T}\mathbf{J}_{i-1}^{\Omega}\mathbf{R}_{i}\cdots & \mathbf{R}_{i,\boldsymbol{q}_{i}}^{T}\mathbf{R}_{i,\boldsymbol{q}_{i}^{T}}\cdots\\
\quad+\mathbf{R}_{i}^{T}\mathbf{J}_{i-1}^{\Omega}\mathbf{R}_{i,\boldsymbol{q}_{i}} & \quad+\mathbf{R}_{i}^{T}\mathbf{R}_{i,\boldsymbol{q}_{i}^{T},\boldsymbol{q}_{i}}
\end{array}\!\!\right]\nonumber \\
 & =\left[\!\!\begin{array}{c|c}
\mathbf{R}_{i}^{T}\mathbf{H}_{i-1}^{\Omega}\mathbf{R}_{i} & \boldsymbol{0}\\
\hline \mathbf{R}_{i,\boldsymbol{q}_{i}}^{T}\mathbf{J}_{i-1}^{\Omega}\mathbf{R}_{i}\cdots & \mathbf{R}_{i,\boldsymbol{q}_{i}}^{T}\mathbf{R}_{i,\boldsymbol{q}_{i}^{T}}\cdots\\
\quad+\mathbf{R}_{i}^{T}\mathbf{J}_{i-1}^{\Omega}\mathbf{R}_{i,\boldsymbol{q}_{i}} & \quad+\mathbf{R}_{i}^{T}\mathbf{R}_{i,\boldsymbol{q}_{i}^{T},\boldsymbol{q}_{i}}
\end{array}\!\!\right]\label{eq:Ho_derivation}
\end{align}

\subsubsection{Linear Body Velocity Hessian\label{a:Hvi_derivation}}

\begin{align}
\mathbf{H}_{i}^{\upsilon} & =\mathbf{J}_{i,\boldsymbol{q}^{i}}^{\upsilon}\nonumber \\
 & =\left(\mathbf{R}_{i}^{T}\left[\begin{array}{c|c}
\mathbf{J}_{i-1}^{\upsilon}+\mathbf{J}_{i-1}^{\Omega}\boldsymbol{p}_{i} & \boldsymbol{p}_{i}\end{array}\right]\right){}_{,\boldsymbol{q}^{i}}\nonumber \\
 & =\left[\!\!\begin{array}{c|c}
\mathbf{R}_{i}^{T}\left(\mathbf{J}_{i-1,\boldsymbol{q}^{i-1}}^{\upsilon}+\mathbf{J}_{i-1,\boldsymbol{q}^{i-1}}^{\Omega}\boldsymbol{p}_{i}\right) & \left(\mathbf{R}_{i}^{T}\boldsymbol{p}_{i,\boldsymbol{q}_{i}^{T}}\right)_{,\boldsymbol{q}^{i-1}}\\
\hline \mathbf{R}_{i,\boldsymbol{q}_{i}}^{T}\left(\mathbf{J}_{i-1}^{\upsilon}+\mathbf{J}_{i-1}^{\Omega}\boldsymbol{p}_{i}\right)\cdots & \mathbf{R}_{i,\boldsymbol{q}_{i}}^{T}\boldsymbol{p}_{i,\boldsymbol{q}_{i}^{T}}\cdots\\
\quad+\mathbf{R}_{i}^{T}\mathbf{J}_{i-1}^{\Omega}\boldsymbol{p}_{i,\boldsymbol{q}_{i}} & \quad+\mathbf{R}_{i}^{T}\boldsymbol{p}_{i,\boldsymbol{q}_{i}^{T},\boldsymbol{q}_{i}}
\end{array}\!\!\right]\nonumber \\
 & =\left[\!\!\begin{array}{c|c}
\mathbf{R}_{i}^{T}\left(\mathbf{H}_{i-1}^{\upsilon}+\mathbf{H}_{i-1}^{\Omega}\boldsymbol{p}_{i}\right) & \boldsymbol{0}\\
\hline \mathbf{R}_{i,\boldsymbol{q}_{i}}^{T}\left(\mathbf{J}_{i-1}^{\upsilon}+\mathbf{J}_{i-1}^{\Omega}\boldsymbol{p}_{i}\right)\cdots & \!\!\!\!\!\!\mathbf{R}_{i,\boldsymbol{q}_{i}}^{T}\boldsymbol{p}_{i,\boldsymbol{q}_{i}^{T}}\cdots\\
\quad+\mathbf{R}_{i}^{T}\mathbf{J}_{i-1}^{\Omega}\boldsymbol{p}_{i,\boldsymbol{q}_{i}} & \quad+\mathbf{R}_{i}^{T}\boldsymbol{p}_{i,\boldsymbol{q}_{i}^{T},\boldsymbol{q}_{i}}
\end{array}\!\!\right]\label{eq:Hv_derivation}
\end{align}

\subsection{Conservative Force Vector, $\left(\boldsymbol{G}_{i}^{g}\right)$\label{a:Conservative-Force-Vector,a}}

\begin{align}
{\boldsymbol{G}_{i}^{g}}^{T} & =m_{i}g^{T}\left(\overline{\boldsymbol{p}}^{i}\right)_{,\left(\boldsymbol{q}^{i}\right)^{T}}\nonumber \\
 & =m_{i}g^{T}{\overline{\mathbf{R}}^{i}}\left\{ {\overline{\mathbf{R}}^{i}}^{T}\left(\overline{\boldsymbol{p}}^{i}\right)_{,\left(\boldsymbol{q}^{i}\right)^{T}}\right\} \nonumber \\
 & =m_{i}g^{T}{\overline{\mathbf{R}}^{i}}\mathbf{J}_{i-1}^{\upsilon}\nonumber \\
 & =m_{i}g^{T}\left({\mathbf{R}^{i-1}}{\overline{\mathbf{R}}_{i}}\right)\mathbf{R}_{i}^{T}\left[\begin{array}{c|c}
\mathbf{J}_{i-1}^{\upsilon}+\mathbf{J}_{i-1}^{\Omega}\boldsymbol{p}_{i} & \mathbf{R}^{i-1}\boldsymbol{p}_{i,\boldsymbol{q}_{i}^{T}}\end{array}\right]\nonumber \\
 & =m_{i}g^{T}{\mathbf{R}^{i-1}}\left({\overline{\mathbf{R}}_{i}}\mathbf{R}_{i}^{T}\right)\left[\begin{array}{c|c}
\mathbf{J}_{i-1}^{\upsilon}+\mathbf{J}_{i-1}^{\Omega}\boldsymbol{p}_{i} & \mathbf{R}^{i-1}\boldsymbol{p}_{i,\boldsymbol{q}_{i}^{T}}\end{array}\right]\nonumber \\
 & =m_{i}g^{T}{\mathbf{R}^{i-1}}\left(\left[\begin{array}{c|c}
\mathbf{J}_{i-1}^{\upsilon}+\mathbf{J}_{i-1}^{\Omega}\overline{\boldsymbol{p}}_{i} & \mathbf{R}^{i-1}\overline{\boldsymbol{p}}_{i,\boldsymbol{q}_{i}^{T}}\end{array}\right]\right)\label{eq:Ggi_derivation}
\end{align}

\bibliographystyle{IEEEtran}
\bibliography{Biblio}

\vspace{-10mm}
\begin{IEEEbiography}[{\includegraphics[width=1in,height=1.25in]{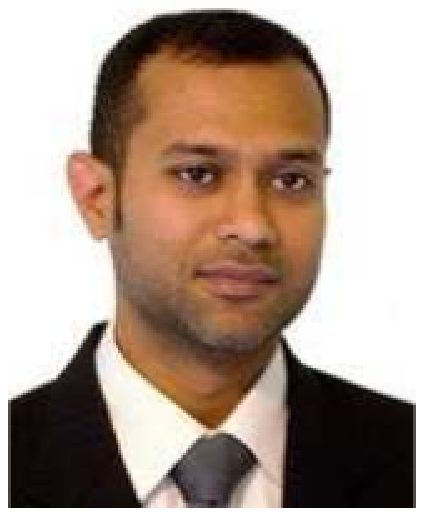}}]{Isuru~S.~Godage}
 (S\textquoteright 10\textendash M\textquoteright 13) is an Assistant
Professor with the School of Computing and the director of the Robotics
and Medical Engineering (RoME) Laboratory at DePaul University. Previously,
he held an Adjoint Assistant Professorship at Vanderbilt University,
and Postdoctoral Research Scholarships at Vanderbilt University and
Clemson University respectively. Dr. Godage holds a Ph.D. in Robotics,
Cognition, and Interaction Technologies from the Italian Institute
of Technology-University of Genoa (2013) and B.Sc. Eng. (Hons) in
Electronic and Telecommunication Engineering from the University of
Moratuwa, Sri Lanka (2007). His research has been funded by National
Science Foundation. Dr. Godage\textquoteright s research interests
include design, modeling, and control of continuum and soft robots
for manipulation and locomotion in applications related to search
and rescue and healthcare applications. 
\end{IEEEbiography}

\vspace{-10mm}

\begin{IEEEbiography}[{\includegraphics[width=1in,height=1.25in]{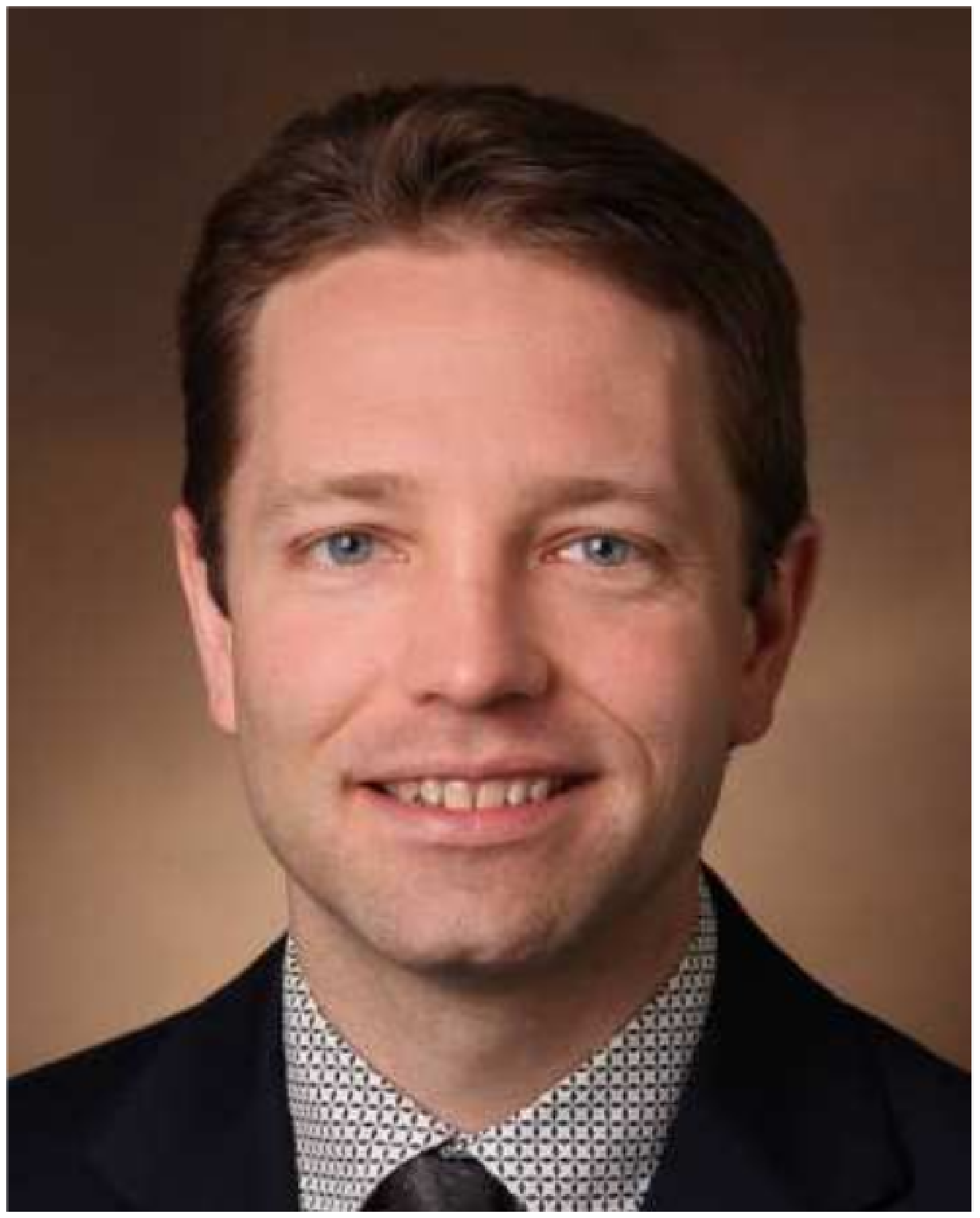}}]{Robert~J.~Webster~III}
 (S\textquoteright 97\textendash M\textquoteright 08\textendash SM\textquoteright 14)
received the B.S. degree in electrical engineering from Clemson University,
Clemson, SC, USA, in 2002, and the M.S. and Ph.D. degrees in mechanical
engineering from Johns Hopkins University, Baltimore, MD, USA, in
2004 and 2007, respectively. In 2008, he joined the Faculty of Vanderbilt
University, Nashville, TN, USA, where he is currently a Professor
of mechanical engineering, electrical engineering, otolaryngology,
neurological surgery, and urologic surgery, and directs the Medical
Engineering and Discovery Laboratory. He is a Member of the Steering
Committee with the Vanderbilt Institute in Surgery and Engineering,
which brings together physicians and engineers to solve challenging
clinical problems. He is a Founder and serves as the President of
Virtuoso Surgical, Inc, Nashville. His current research interests
include surgical robotics, image-guided surgery, and continuum robotics.
Dr. Webster was the recipient of the IEEE Robotics and Automation
Society Early Career Award, the National Science Foundation CAREER
Award, the Robotics Science and Systems Early Career Spotlight Award,
the IEEE Volz Award, and the Vanderbilt Engineering Award for Excellence
in Teaching. He is the Chair of the International Society for Optics
and Photonics Image-Guided Procedures, Robotic Interventions, and
Modeling Conference.
\end{IEEEbiography}

\vspace{-10mm}

\begin{IEEEbiography}[{\includegraphics[bb=0bp 0bp 244bp 305bp,width=1in,height=1.25in]{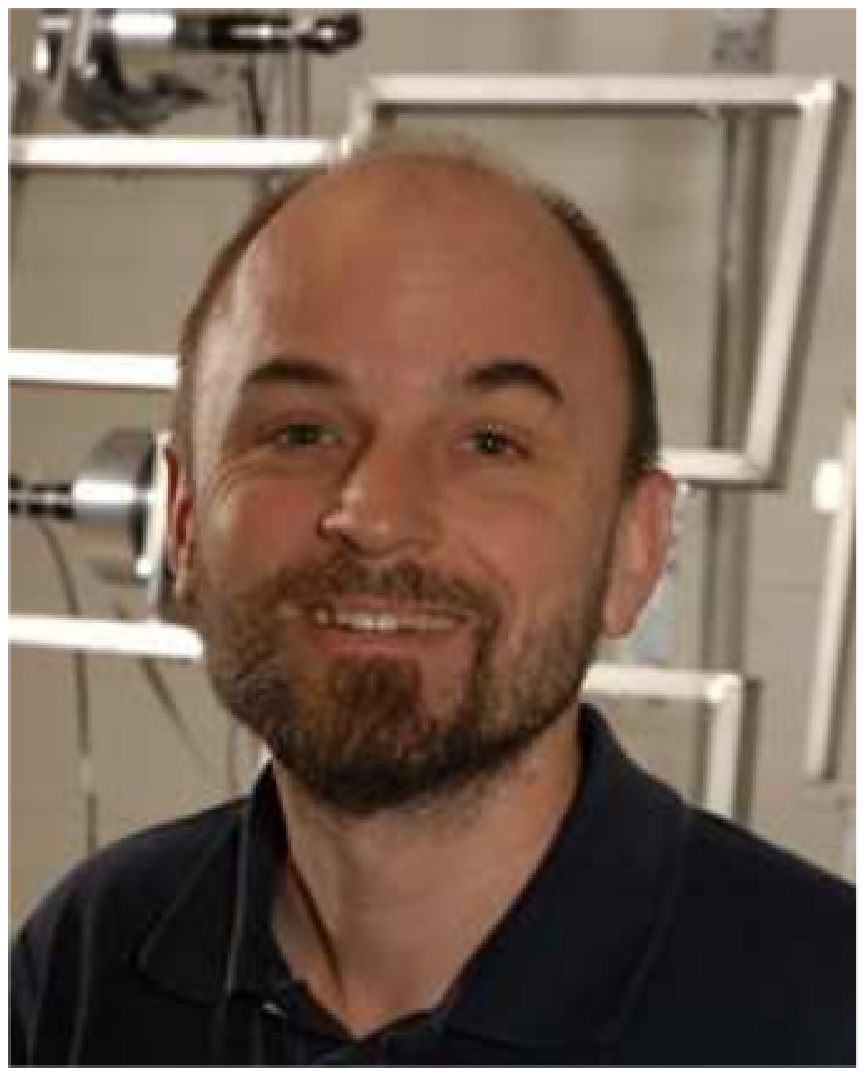}}]{Ian~D.~Walker}
 (S\textquoteright 84\textendash M\textquoteright 85\textendash SM\textquoteright 02\textendash F\textquoteright 06)
received the B.Sc. degree in Mathematics from the University of Hull,
UK, in 1983, and the M.S. and Ph.D. degrees, both in Electrical Engineering,
in 1985 and 1989, respectively, from the University of Texas at Austin.
He has served as Vice President for Financial Activities for the IEEE
Robotics and Automation Society, and as Chair of the AIAA Technical
Committee on Space Automation and Robotics. He has also served on
the Editorial Boards of the IEEE Transactions on Robotics, the IEEE
Transactions on Robotics and Automation, the International Journal
of Robotics and Automation, the IEEE Robotics and Automation Magazine,
and the International Journal of Environmentally Conscious Design
and Manufacturing. He currently serves on the Editorial Board of Soft
Robotics. Dr. Walker's research interests include biologically inspired
and continuum robotics, as well as architectural robotics. 

\end{IEEEbiography}

\end{document}